\documentclass[journal]{IEEEtran}
\IEEEoverridecommandlockouts

\usepackage{cite}
\usepackage{amsmath,amssymb,amsfonts}
\usepackage{algorithmic}
\usepackage{graphicx}
\usepackage{subcaption}
\usepackage{textcomp}
\usepackage{xcolor}
\usepackage{booktabs}
\usepackage{float}
\usepackage{makecell}
\usepackage{multirow}
\usepackage{tabularx}
\usepackage{xr-hyper}
\makeatletter
\newcommand*{\addFileDependency}[1]{%
  \typeout{(#1)}%
  \@addtofilelist{#1}%
  \IfFileExists{#1}{}{\typeout{No file #1.}}%
}
\makeatother
\newcommand*{\myexternaldocument}[1]{%
  \externaldocument{#1}%
  \addFileDependency{#1.tex}%
  \addFileDependency{#1.aux}%
}
\myexternaldocument{appendix}
\usepackage[hidelinks]{hyperref}

\def\BibTeX{{\rm B\kern-.05em{\sc i\kern-.025em b}\kern-.08em
    T\kern-.1667em\lower.7ex\hbox{E}\kern-.125emX}}
\begin{document}

\title{
  DiPLO: Self-distilled \underline{Di}ffusion \underline{P}lanning for \underline{Lo}co-manipulation 
}

\author{%
\IEEEauthorblockN{Shafeef Omar\textsuperscript{1,4},
Sebastian Sanokowski\textsuperscript{1},
Elham Daneshmand\textsuperscript{2},
Dian Yu\textsuperscript{1}}
\IEEEauthorblockN{Daniel Cremers\textsuperscript{1},
Cordelia Schmid\textsuperscript{3,4},
Majid Khadiv\textsuperscript{1}
}

\thanks{\textsuperscript{1}Technical University of Munich (TUM), Munich, Germany.}
%
\thanks{\textsuperscript{2}McGill University, Montreal, Canada.}
%
\thanks{\textsuperscript{3}INRIA, Paris, France.}
%
\thanks{\textsuperscript{4}Institute for Advanced Study (IAS), Technical University of Munich, Munich, Germany.}
}
\makeatletter
\let\@titleblock\@maketitle
\def\@maketitle{\@titleblock
    \vskip 3.0\baselineskip
    \noindent\begin{minipage}{\textwidth}
        \centering
        \setcounter{figure}{0}%
        \includegraphics[width=\textwidth]{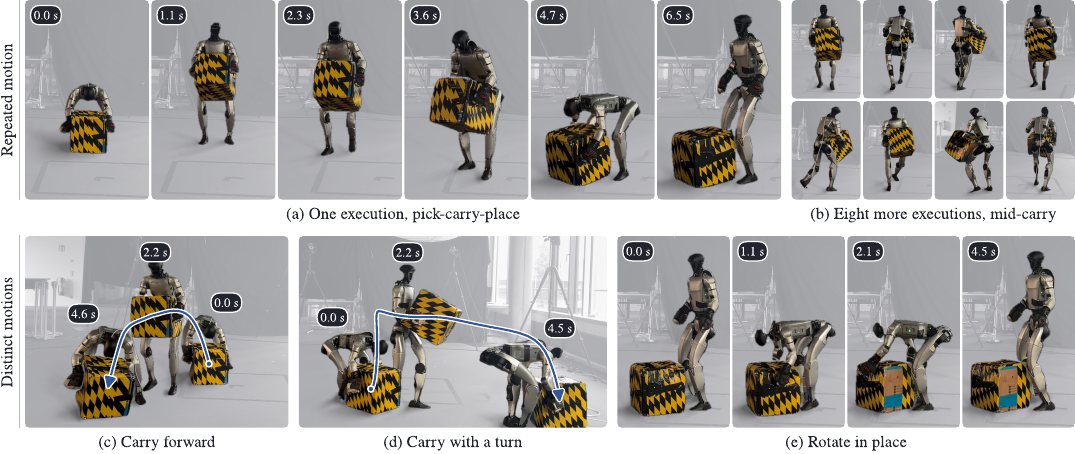}
        \captionof{figure}{
        Real-robot executions of our planner-tracker architecture for humanoid loco-manipulation on the Unitree G1. 
        }
        \label{fig:real_robot}
    \end{minipage}\par\vspace{-1.78\baselineskip}}
\makeatother
\maketitle

\begin{figure*}[t]
    \centering
    \includegraphics[width=\textwidth]{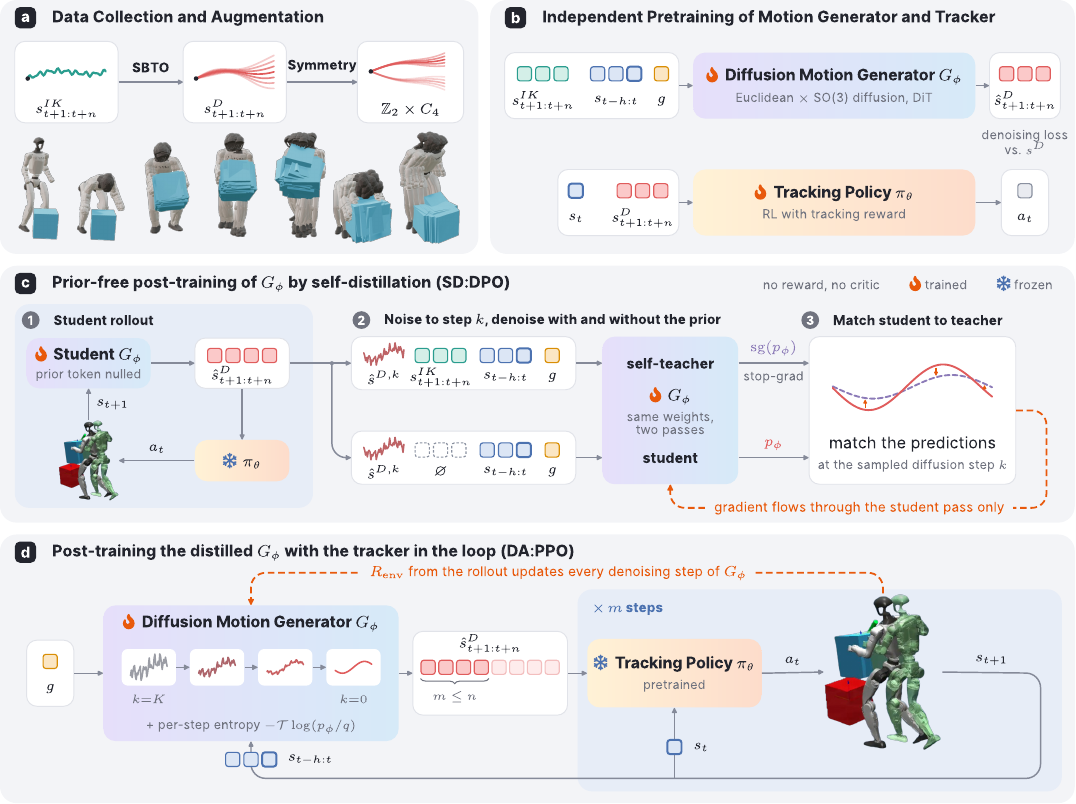}
    \caption{
    Overview of the pipeline. (a) SBTO turns each IK-retargeted demonstration $s^{IK}$ into a set of dynamically feasible motions $s^{D}$, and symmetries further enlarge the data corpus. (b) The diffusion motion generator $G_\phi$ is pretrained to generate an $n$-step feasible motion conditioned on the IK-retargeted motion prior, the state history and the goal, while the tracking policy $\pi_\theta$ is trained separately with RL to track $s^{D}$. (c) Self-distillation with SD:DPO removes the need for the task prior: the student $G_\phi$, with the prior token nulled, rolls out with the frozen tracker $\pi_\theta$; each plan it produces is noised to a random denoising step $k$ and denoised twice by the same network, with and without $s^{IK}$, and the student's reverse kernel $p_\phi$ is matched to the self-teacher's stop-gradient kernel $\mathrm{sg}(p_\phi)$. (d) The distilled $G_\phi$ is then post-trained with DA:PPO, conditioned on the state history and the goal alone. In closed loop, $G_\phi$ plans $n$ steps, $\pi_\theta$ executes the first $m \leq n$ of them, and $G_\phi$ re-plans from the reached state. DA:PPO credits the rollout reward $R_{\mathrm{env}}$ to every denoising step of $G_\phi$ while $\pi_\theta$ stays fixed; the deployed pipeline runs the same loop without the reward.
    }
    \label{fig:pipeline}
\end{figure*}
\begin{abstract}


Humanoid loco-manipulation is particularly challenging because it requires coordinating locomotion, balance, and object interaction, making it difficult to learn from limited demonstrations or from scratch through Reinforcement Learning (RL). We introduce DiPLO (self-distilled Diffusion Planning for Loco-manipulation), a framework that combines a self-distilled diffusion planner with a low-level whole-body tracking policy. Our planner is pretrained via supervised learning on motions derived from human-to-humanoid retargeting, while the tracker is trained via RL to execute them. DiPLO centers around two complementary post-training stages. First, we introduce Self-Distillation Diffusion Policy Optimisation (SD:DPO) to remove the planner’s dependence on strong task conditioning, which we identify as a crucial ingredient for effective pretraining, as downstream fine-tuning does not recover comparable performance when it is omitted.. We then fine-tune this distilled model with Diffusion-Augmented Proximal Policy Optimization (DA:PPO), a maximum-entropy RL method for diffusion models, to improve task success and align generated plans with the frozen tracker’s execution capabilities. Together, these stages enable motion generation through state-history and goal conditioning alone, enabling $\sim99\%$ task success in pick-carry-place tasks, compared to just $45\%$ for the pretrained model. The resulting planner also generates motions for unseen object goals, with further evaluations on kicking and pushing/pulling tasks. Finally, we deploy our pipeline on the hardware, using our DA:PPO planner-tracker on a Unitree G1, achieving repeated successful executions without any real-world fine-tuning. Video: \url{https://youtu.be/rDrdnYQUYoM}

\end{abstract}


\section{Introduction}
Humanoid loco-manipulation requires coordinating locomotion, balance, and object interaction. Human motion data provides useful supervision for learning these behaviors, but collecting large datasets is costly~\cite{khazatsky2024droid, li2023omomo}. Our goal is to learn from a small dataset of retargeted motions while keeping task specification simple: the user provides the desired object position and orientation, and the robot determines how to reach it. This requires a planner that can generate trackable motions for different object goals, including goals beyond those represented in the demonstrated behaviours.


We introduce DiPLO (self-distilled Diffusion Planning for Loco-manipulation), which combines a diffusion motion planner with an RL-trained tracker. Starting from 29 retargeted robot--object motion sequences from an existing dataset~\cite{yang2025omniretargetinteractionpreservingdatageneration} (which amounts to just $2.7$ mins of data), we use trajectory optimization~\cite{dhedin2026dynaretargetdynamicallyfeasibleretargetingusing} and augmentation to generate dynamically feasible training motions for both components. During execution, the diffusion planner generates motion chunks and replans from the measured state as the tracker executes them. 
Such a pretrained planner still relies strongly on kinematic reference conditioning, which limits its ability to generate motions to new goals.
To remove this dependence after pretraining, we propose Self-Distillation Diffusion Policy Optimization (SD:DPO). SD:DPO transfers guidance from a reference-conditioned teacher to a student conditioned only on state history and the object goal. The student learns to match the teacher's denoising predictions at states encountered during environment rollouts with the tracker. These rollouts expose the student to the consequences of planning and tracking errors, extending its supervision beyond the contexts represented in the training data. SD:DPO thereby enables high survival without requiring a kinematic reference at deployment.

Trackable motions alone, however, do not ensure that the object reaches its goal. We therefore follow self-distillation with planner fine-tuning using DA:PPO~\cite{sanokowski2025damdp}, a maximum-entropy RL method for diffusion models. With the tracker fixed, DA:PPO improves task completion through closed-loop interaction while encouraging diversity in generated plans. SD:DPO and DA:PPO thus address complementary requirements: removing dependence on reference conditioning and improving task success.
Our main contributions are:
\begin{itemize}
    \item We introduce DiPLO, a framework that combines a self-distilled diffusion-based planner for humanoid loco-manipulation that jointly generates whole-body and object motion for execution by a single generalist RL tracking policy.

    \item We propose SD:DPO, which distills a reference-conditioned diffusion planner into a goal-conditioned planner, removing the need for kinematic reference conditioning at deployment. We further apply DA:PPO fine-tuning through closed-loop interaction with a fixed genralist tracker, improving object-goal completion while maintaining diverse, trackable plans.

    \item We provide extensive ablations across a diverse set of loco-manipulation tasks, demonstrating the effectiveness of DiPLO and its components compared with state-of-the-art baselines, together with zero-shot transfer to a real humanoid robot.
\end{itemize}



\section{Related Work}

\subsubsection{Learning from retargeted motion data}
Retargeted motions provide supervision for learning whole-body tracking policies~\cite{2018-TOG-deepMimic, luo2023phc, he2024omnih2o}, but kinematic retargeting~\cite{ze2025gmr, yang2025omniretargetinteractionpreservingdatageneration} alone does not ensure physically feasible robot--object interaction. DynaRetarget~\cite{dhedin2026dynaretargetdynamicallyfeasibleretargetingusing} addresses this limitation through sampling-based trajectory optimization (SBTO), which refines a kinematic reference into a dynamically feasible motion. To scale demonstrations beyond those recorded, HumanoidMimicGen~\cite{lin2026humanoidmimicgen} augments loco-manipulation data by adapting object-centric skill segments of a source demonstration to new object poses. However, it plans kinematically and decouples locomotion from stationary manipulation, so dynamic feasibility is not enforced during generation. In contrast, we repurpose SBTO, originally proposed to refine individual demonstrations, for data augmentation: starting from perturbed initial conditions, we exploit the stochasticity of the sampling-based optimizer to synthesize locally diverse, dynamically feasible robot--object motions in which locomotion and manipulation remain coupled.

\subsubsection{Diffusion planning for loco-manipulation}
Diffusion Policy~\cite{chi2024diffusionpolicyvisuomotorpolicy} models multimodal action sequences for robotic manipulation. In humanoid locomotion, however, maintaining balance while accomplishing the task motivates delegating execution to a low-level controller \cite{ha2025learning}. PDP~\cite{truong2024pdp} learns action generation from expert policies, while BeyondMimic~\cite{liao2025beyondmimicmotiontrackingversatile} uses guided state--action diffusion to generate goal-directed behaviors. An alternative is to separate motion generation from execution: ~\cite{zhang2026learning} combine a diffusion motion generator with a tracking policy for terrain locomotion.
Recent works have shown the success of this separation in loco-manipulation settings \cite{gu2026refinedp,debbad2026humanoid}. 
REFINE-DP~\cite{gu2026refinedp} is closest to our approach in combining diffusion planning with RL fine-tuning for humanoid loco-manipulation. Its planner generates motion commands for separate upper- and lower-body controllers. DiPLO instead jointly plans whole-body and object motion for a single generalist tracking policy.
Our post-training procedure also differs. We first use SD:DPO to remove the dependence on kinematic reference conditioning introduced during pretraining on retargeted motions. We then apply maximum-entropy fine-tuning with DA:PPO while keeping the tracker fixed. Finally, our generalization evaluation examines whether the same trained planner can reach different requested object positions and orientations that are absent from the retargeted dataset. 

Our planner also treats rotational features through diffusion on $\mathrm{SO}(3)$. Such processes have been developed for rotational alignment~\cite{leach2022denoising, jagvaral2024unified}, while DiffDock~\cite{corso2023diffdock} combines diffusion over translational, rotational, and torsional degrees of freedom. We apply this geometric formulation to humanoid--object motion, evaluate its effect on closed-loop survival, and retain it during planner post-training.

\subsubsection{Distillation in Robot Learning}
Teacher--student distillation allows policies to benefit from information unavailable at deployment \cite{lee2020quadruped}. Following such a recipe, OmniH2O~\cite{he2024omnih2o} distills a controller with privileged observations into a policy using deployable observations and sparse motion commands. Our objective concerns the high-level task specification: removing the diffusion planner's dependence on strong task conditioning (kinematic reference trajectory).

The closest methodological precedent is Self-Distillation Policy Optimization (SDPO)~\cite{hubotter2026reinforcement}, which uses a language model conditioned on additional feedback as a self-teacher and distills its next-token predictions into a student. Inspired by SDPO, we propose SD:DPO for diffusion motion planning. A teacher with access to privileged task-prior conditioning supervises the student's denoising predictions on states encountered during closed-loop environment rollouts with the tracker, enabling reward-free credit assignment through denoising-step matching. This distills the guidance provided by the privileged task prior into a planner conditioned only on the robot's state history and object goal, allowing the task prior to be removed at deployment. We evaluate both the resulting improvement in survival and the ability of the distilled planner to generate motions for object goals absent from the retargeted dataset.

\subsubsection{Fine-tuning of diffusion planners}
Diffusion-Policy Policy Optimization (DPPO)~\cite{dppo2024} enables reinforcement learning for diffusion policies by treating denoising transitions as decisions with tractable likelihoods and applying PPO to the resulting augmented process. REFINE-DP~\cite{gu2026refinedp} applies diffusion-policy fine-tuning to humanoid loco-manipulation, considering both a fixed controller and alternating updates of the planner and controllers. However, DPPO promotes exploration by truncating the denoising process during training, which can lead to jittery motions. Such behavior is particularly undesirable for real-world humanoid control, where smooth and dynamically consistent motions are essential.

Diffusion-Augmented MDPs~\cite{sanokowski2025damdp} instead provide a maximum-entropy formulation in which ratios of reverse and forward transition densities yield tractable entropy regularization at each denoising step. We use their PPO instantiation, DA:PPO, to fine-tune the planner after SD:DPO. Our setting combines Euclidean and $\mathrm{SO}(3)$ diffusion and evaluates generated motions through a frozen generalist tracker. DA:PPO optimizes task completion after kinematic reference conditioning has been removed, while entropy regularization encourages diversity in the generated plans. We separately evaluate tracker adaptation to distinguish improvements due to planner fine-tuning from those obtained by adapting the tracker.


\section{Preliminaries}
\label{sec:preliminaries}

\subsection{Motion Planning and Tracking}
\label{subsec:planner_tracker}

We consider humanoid loco-manipulation with a high-level motion planner and a low-level tracking policy. The planner generates desired whole-body and object motion over a finite horizon, while the tracker executes this motion by adjusting joint actions based on the measured robot and object states. This hierarchical decomposition is motivated by the underactuated nature of humanoid robots and the need to maintain balance while tracking task-level motions.

Let $s_t$ denote the measured robot--object motion state and $g_t$ the desired object pose expressed relative to its current pose. At planning step $t$, the planner $G_\phi$ receives the state history $s_{t-h:t}$ over $h$ steps and the goal $g_t$. It may additionally receive a kinematic reference $s^{IK}_\tau$ obtained by inverse-kinematics (IK) retargeting. These inputs condition the generation of an $n$-frame motion chunk over $\tau=t+1:t+n$:
{
\begin{equation}
\label{eq:planner_interface}
\begin{aligned}
c_t&=(s_{t-h:t},g_t),\\
\hat s^D_\tau&\sim G_\phi(\cdot\mid c_t,s^{IK}_\tau).
\end{aligned}
\end{equation}
}
Here, $\hat s^D_\tau$ denotes a generated motion chunk whereas $s^D_\tau$ denotes a motion chunk from the dataset. When the kinematic reference is unavailable, its input is replaced by a null token $\varnothing$. We use $z_\tau$ to collect the planner's conditioning: $z_\tau=(c_t,s^{IK}_\tau)$ with the reference and $z_\tau=c_t$ without it. Section~\ref{subsec:motion_representation} specifies the motion state $s_t$.

The generated chunk provides the reference motion for the tracker. At control step $j$, the policy $\pi_\theta$ receives robot and object observations $o_j$, its previous action $a_{j-1}$, and planned motion features $\mathcal R_j(\hat s^D_\tau)$. Here $\mathcal R_j$ extracts reference robot and object states at specified future offsets from $j$. The policy predicts a residual correction to the reference joint positions $q_j^{\mathrm{ref}}$. Denoting the scaled residual by $a_j$, the joint controller's target is
{
\begin{equation}
\label{eq:tracker_interface}
\begin{aligned}
a_j&\sim\pi_\theta\!\left(\cdot\mid o_j,a_{j-1},
\mathcal R_j(\hat s^D_\tau)\right),\\
q_j^{\mathrm{PD}}&=q_j^{\mathrm{ref}}+a_j.
\end{aligned}
\end{equation}
}
A proportional--derivative (PD) controller uses $q_j^{\mathrm{PD}}$ as its joint-position target. The tracker executes the first $m\leq n$ frames of the chunk, at control steps $j=t,\ldots,t+m-1$. The planner then generates a new chunk from the updated measured history. Replanning thus accounts for the motion actually executed, including tracking errors. 

To train the tracker, reference motions are sampled from a dataset $\mathcal D$ and executed in simulation~\cite{2018-TOG-deepMimic,liao2025beyondmimicmotiontrackingversatile}. We train a policy via PPO~\cite{schulman2017ppo} to learn to follow these motions by maximizing the expected tracking return:
{
\begin{equation}
\label{eq:tracker_objective}
J_{\mathrm{track}}(\theta)
=\mathbb E_{s^D\sim\mathcal D,\,\pi_\theta}
\!\left[\sum_{j=0}^{T-1}\gamma_{\mathrm{track}}^j
r_{\mathrm{track},j}\right].
\end{equation}
}
Where the expectation is over sampled references and simulated rollouts over the whole episode length $T$, and $\gamma_{\mathrm{track}}$ discounts the control steps. The reward $r_{\mathrm{track},j}$ measures how closely the executed robot and object motions follow the reference, with additional contact and regularization terms. During this training stage, the reference comes from the dataset; during planner execution, it comes from the generated chunk.

\subsection{Conditional Diffusion Models}
\label{subsubsec:diffusion_processes}

We use a conditional diffusion model as a planner to learn to generate motion by reversing a prescribed noising process~\cite{sohldickstein2015deep,ho2020denoisingdiffusionprobabilisticmodels}. For training, a clean chunk $s^{D,0}_\tau=s^D_\tau$ and its conditioning $z_\tau$ are drawn from the dataset. The forward transitions $s^{D,k}_\tau\sim q^k(\cdot\mid s^{D,k-1}_\tau,z_\tau)$ progressively corrupt the motion over $K$ diffusion levels. At the final level, the distribution approximately matches a prescribed diffusion prior $p(\cdot\mid z_\tau)$. This prior supplies the initial noise for the denoising process.

Learned reverse transitions then progressively denoise it using the conditioning $z_\tau$:
{
\begin{equation}
\label{eq:diffusion_generation}
\begin{aligned}
\hat s^{D,K}_\tau&\sim p(\cdot\mid z_\tau),\\
\hat s^{D,k-1}_\tau&\sim p_\phi\!\left(
\cdot\mid\hat s^{D,k}_\tau,z_\tau\right),
\quad k=K,\ldots,1.
\end{aligned}
\end{equation}
}
The complete chain defines the planner $G_\phi$; each kernel $p_\phi$ describes one denoising step which is parameterized by a neural network which receives the noisy chunk, diffusion level $k$, and conditioning $z_\tau$. The final sample $\hat s^{D,0}_\tau=\hat s^D_\tau$ is passed to the tracking policy. Pretraining uses dataset chunks corrupted at sampled noise levels to construct supervised targets for the network. See Section~\ref{subsec:diffusion_pretraining} for more details.

\subsection{Diffusion-Augmented MDPs and DA:PPO}
\label{subsec:damdp_preliminaries}

The planner receives an environment reward when the tracker executes the first $m$ frames of a generated chunk. To improve the planner with RL, this reward must guide the denoising decisions that produced the chunk. A Diffusion-Augmented Markov Decision Process (DA-MDP)~\cite{sanokowski2025damdp} makes these decisions explicit: the diffusion augmented state $\tilde s_k$ contains the planner's conditioning $z$, the current noisy chunk $\tilde a^k$, and its diffusion level $k$; the action is the next, less noisy chunk $\tilde a^{k-1}$. Writing $\tilde a^k=\hat s^{D,k}_\tau$ and suppressing the window subscript $\tau$, we obtain
{
\begin{equation}
\label{eq:damdp_state_action}
\tilde s_k=(z,\tilde a^k,k),
\qquad
\tilde a^{k-1}\sim p_\phi(\cdot\mid\tilde s_k).
\end{equation}
}

Each action has the tractable density $p_\phi(\tilde a^{k-1}\mid\tilde s_k)$, even though the marginal density of the completed chunk is intractable. For $k>1$, the environment remains unchanged and the next state contains the updated noisy chunk at level $k-1$. At $k=1$, the tracker executes the first $m$ frames, the planner's conditioning is updated, and a fresh prior sample starts the next denoising chain. For a successor state $\tilde s'=(z',a',k')$, these two cases give
{
\begin{equation}
\label{eq:damdp_transition}
\begin{aligned}
&p_{\mathrm{DA}}(\tilde s'\mid\tilde s_k,\tilde a^{k-1})\\
&=\begin{cases}
\delta_z(z')\,\delta_{\tilde a^{k-1}}(a')\,\delta_{k-1}(k'), & k>1,\\
P_m(z'\mid z,\tilde a^0)\,p(a'\mid z')\,\delta_K(k'), & k=1.
\end{cases}
\end{aligned}
\end{equation}
}
Here $\delta_x$ denotes a point mass at $x$, and $P_m$ is the transition induced by executing the first $m$ frames with a fixed tracker. The resulting observations update the history and goal direction in $z'$ and optionally the kinematic reference. Only the final denoising decision receives the reward $r$ accumulated during execution, so $\tilde r_k=\mathbf 1_{\{k=1\}}r$. The discount likewise applies only when execution advances the environment: $\bar\gamma_k=\gamma$ for $k=1$ and one otherwise, where $\gamma$ discounts planning steps.

The maximum-entropy formulation also regularizes the denoising decisions. Its action-value function $Q_{\mathrm{DA}}$ and state-value function $V_{\mathrm{DA}}$ satisfy the soft Bellman equations
\begin{equation}
\label{eq:damdp_bellman}
\begin{aligned}
Q_{\mathrm{DA}}(\tilde s,\tilde a)
&=\mathbb E\!\left[\tilde r+\bar\gamma_k V_{\mathrm{DA}}(\tilde s')
\mid\tilde s,\tilde a\right],\\
V_{\mathrm{DA}}(\tilde s)
&=\mathbb E_{\tilde a\sim p_\phi(\cdot\mid\tilde s)}
\!\left[Q_{\mathrm{DA}}(\tilde s,\tilde a)-\mathcal T\ell_\phi(\tilde s,\tilde a)\right],
\end{aligned}
\end{equation}
with zero continuation value at terminal states. $\ell_\phi(\tilde s_k,\tilde a^{k-1})=\log\!\big[p_\phi(\tilde a^{k-1}\mid\tilde s_k)/q_\phi^k(\tilde a^k\mid\tilde a^{k-1},z)\big]$ is a diffusion regularization term and given by the log ratio of the reverse transition density to the corresponding forward density. It replaces the action log-likelihood in the usual soft Bellman equations~\cite{haarnoja2018soft}. Thus, $Q_{\mathrm{DA}}$ combines the current environment reward with the value of the next augmented state, while $V_{\mathrm{DA}}$ also includes the current diffusion regularization term. These recursions propagate the reward from execution through all preceding denoising decisions.
$-\mathcal T\ell_\phi$ encourages entropy in the reverse process while favoring denoising paths that are plausible under forward noising. Retaining this diversity during fine-tuning encourages exploration whose temperature $\mathcal T$ controls the strength of the regularization.

DA:PPO applies PPO~\cite{schulman2017ppo} to these individual denoising decisions. Let $i$ index transitions collected under $p_{\phi_{\mathrm{old}}}$, and let the critic $V_\psi$ approximate $V_{\mathrm{DA}}$. Generalized advantage estimation~\cite{schulman2016gae} computes $A_i^{\mathrm{old}}$ from the residuals $\delta_i=\tilde r_i-\mathcal T\ell_i^{\mathrm{old}}+\bar\gamma_iV_\psi(\tilde s_{i+1})-V_\psi(\tilde s_i)$, where $\ell_i^{\mathrm{old}}=\ell_{\phi_{\mathrm{old}}}(\tilde s_i,\tilde a_i)$. The advantage therefore accounts for both execution rewards and the log-ratio contributions along the rollout.

Because the log ratio changes as the policy is updated, DA:PPO replaces its current-step contribution in the stored advantage: $\widehat A_i(\phi)=\operatorname{sg}\!\left[A_i^{\mathrm{old}}-\mathcal T(\ell_i(\phi)-\ell_i^{\mathrm{old}})\right]$. The stop-gradient operator $\operatorname{sg}$ holds this quantity fixed when differentiating the policy loss. With the reverse-kernel importance ratio $\rho_i=p_\phi(\tilde a_i\mid\tilde s_i)/p_{\phi_{\mathrm{old}}}(\tilde a_i\mid\tilde s_i)$, the loss is
\begin{equation}
\label{eq:dappo_loss}
\begin{aligned}
\mathcal L(\phi)={}&\mathbb E_i\big[
-\min\!\big(\rho_i\widehat A_i,\,
\mathrm{clip}(\rho_i,1-\epsilon,1+\epsilon)\widehat A_i\big)\\
&\hspace{7mm}-\mathcal T\operatorname{sg}(\chi_i\rho_i)\log q_i(\phi)\big].
\end{aligned}
\end{equation}
The first term is the clipped PPO objective, with clipping range set by $\epsilon$. The second accounts for optionally learned diffusion coefficients in the forward density $q_i(\phi)$, evaluated on transition $i$; it contributes no gradient if diffusion coefficients are fixed. The mask $\chi_i$ applies this correction only where the PPO objective is not clipped: it is zero when $\widehat A_i\geq0$ and $\rho_i>1+\epsilon$, or when $\widehat A_i<0$ and $\rho_i<1-\epsilon$, and one otherwise.

\section{Method}
\label{sec:method}

DiPLO learns a diffusion motion planner and a single generalist tracking policy from dynamically feasible motions constructed from retargeted data. After pretraining, we hold the tracker fixed and improve the planner in two stages. SD:DPO removes its dependence on kinematic reference conditioning; subsequent DA:PPO fine-tuning optimizes towards better task completion through closed-loop execution.

\subsection{Robot--Object Motion Representation}
\label{subsec:motion_representation}

The planner represents observed and generated motion through the same robot--object state:
\begin{equation}
\label{eq:state}
s_t=\big(
\underbrace{\Delta p^{xy}_t,\ \Delta\psi_t,\ h_t,\ R_t,\ q_t}_{\text{robot motion}},
\underbrace{p^o_t,\ R^o_t}_{\text{object motion}}\big).
\end{equation}
The robot states comprise planar root displacement $\Delta p^{xy}_t\in\mathbb R^2$, heading change $\Delta\psi_t\in\mathbb R$, root height $h_t$, heading-invariant root orientation $R_t\in\mathrm{SO}(3)$, and joint positions $q_t\in\mathbb R^{29}$. The object states are its position $p^o_t\in\mathbb R^3$ and orientation $R^o_t\in\mathrm{SO}(3)$. Planar displacements and object orientations are expressed in the current robot heading frame, while heights remain absolute. The representation therefore removes dependence on global planar position and heading. At the network input, each rotation is encoded in the continuous 6D representation, giving 48 features per time-frame.

\subsection{Pretraining from Retargeted Motions}
\label{subsec:diffusion_pretraining}

\subsubsection{Training motions}
\label{subsec:data_collection}

Kinematic retargeting provides motion references but does not enforce the robot's dynamics or contact constraints. We refine these references with sampling-based trajectory optimization (SBTO)~\cite{dhedin2026dynaretargetdynamicallyfeasibleretargetingusing}. By perturbing the initial conditions and varying the optimization seeds we generate several dynamically feasible variations of each motion. Additionally, we expand this dataset by mirroring robot and object motions across the sagittal plane and applying the box's quarter-turn rotational symmetries (Fig.~\ref{fig:SBTO_aug}). Each training pair contains a kinematic reference $s^{IK}_\tau$ and a feasible target motion $s^D_\tau$.


\subsubsection{Motion planner}
\label{subsubsec:motion_planner}

We use a Diffusion Transformer (DiT)~\cite{peebles2023dit} as the denoiser. It receives the noisy chunk, diffusion level $k$, state history and goal $c_t$, and kinematic reference $s^{IK}_\tau$. During pretraining, we replace the reference with the null token $\varnothing$ in 50\% of examples, so the network learns to denoise with either form of conditioning. Appendix~\ref{app:architecture} provides additional architecture details.

We use separate diffusion processes for Euclidean features and rotations. Grouping these features across the chunk, we write $s^{D,k}_\tau=(x^k,R^k)$ for noised training data and $\hat s^{D,k}_\tau=(\hat x^k,\hat R^k)$ for a generated sample. Conditioning is omitted below for readability.

\textbf{Euclidean features.} After normalization, the Euclidean features follow discrete time variance-preserving (VP) diffusion with a cosine noise schedule~\cite{nichol2021improved}. Each forward step scales down the signal and adds Gaussian noise:
\begin{equation}
\label{eq:euc_forward}
q(x^k\mid x^{k-1})
=\mathcal N\!\left(\sqrt{\alpha_k}\,x^{k-1},\,\beta_k I\right),
\end{equation}
where $\alpha_k=1-\beta_k$ and $\bar\alpha_k=\prod_{j=1}^k\alpha_j$, with $\bar\alpha_0=1$. This choice preserves unit marginal variance for initially normalized features. For training, we sample any noise level directly as $x^k=\sqrt{\bar\alpha_k}x^0+\sqrt{1-\bar\alpha_k}\epsilon$, where $\epsilon\sim\mathcal N(0,I)$. As $\bar\alpha_K$ approaches zero, the terminal distribution approaches the diffusion prior $\mathcal N(0,I)$.

Denoising requires the optimal score $u_k^*(s^{D,k}_\tau)=\nabla_{x^k}\log q_k(x^k,R^k)$, which describes how the noisy data density changes with the Euclidean features~\cite{song2021scorebased}. A learned estimate $u_{\phi,k}$ determines the mean of the DDPM reverse kernel:
{
\begin{equation}
\label{eq:euc_reverse}
\begin{aligned}
p_\phi(\hat x^{k-1}\mid\hat s^{D,k}_\tau)
&=\mathcal N(\mu_\phi^k,\tilde\beta_k I),\\
\mu_\phi^k
&=\frac{\hat x^k+\beta_k u_{\phi,k}(\hat s^{D,k}_\tau)}{\sqrt{\alpha_k}},
\end{aligned}
\end{equation}
}
where $\tilde\beta_k=\beta_k(1-\bar\alpha_{k-1})/(1-\bar\alpha_k)$. The score term provides the denoising correction, and the variance controls the remaining noise. This Gaussian kernel approximates the finite-step reverse distribution.

Although the data score is unknown, the clean sample and sampled noise provide a computable training target. We use $\mathbf v$-prediction~\cite{salimans2022progressive}, with target $\mathbf v^k=\sqrt{\bar\alpha_k}\epsilon-\sqrt{1-\bar\alpha_k}x^0$. The network output $\hat{\mathbf v}_\phi^k$ is converted to a score estimate by
\begin{equation}
\label{eq:v_to_score}
u_{\phi,k}(\hat s^{D,k}_\tau)
=-\hat x^k-\sqrt{\frac{\bar\alpha_k}{1-\bar\alpha_k}}\,\hat{\mathbf v}_\phi^k.
\end{equation}
Substituting this estimate into~\eqref{eq:euc_reverse} gives the mean used for denoising. With squared-error regression, the optimal predictor is the conditional mean of $\mathbf v^k$, and this conversion recovers the data score at noise level $k$. We use smooth-L1 regression in practice (see  Appendix~\ref{app:pretrain_hparams} for more details).

\textbf{Rotational features.} We use a discrete time analogue of VP diffusion on $\mathrm{SO}(3)$, contracting root and object rotations towards a reference orientation $A\in\mathrm{SO}(3)$ while increasing rotational noise. We set $A=I$ for the root and use the current measured object orientation for the object. Each reference remains fixed throughout the chunk's diffusion process.

To express these operations, let $v=(v_1,v_2,v_3)^\top\in\mathbb R^3$ be a rotation vector whose direction defines the rotation axis and whose norm is the rotation angle. The map $v\mapsto[v]_\times\in\mathfrak{so}(3)$ forms the corresponding skew-symmetric matrix, and its corresponding rotation matrix can be computed with $\mathrm{Exp}(v)=\exp([v]_\times)\in\mathrm{SO}(3)$. Conversely, $\mathrm{Log}(R)\in\mathbb R^3$ inverts this operation back to its rotation vector and the corresponding matrix logarithm is defined as $\log R=[\mathrm{Log}(R)]_\times$.

We model rotational noise with the isotropic $\mathcal{IG}_{\mathrm{SO}(3)}$ distribution~\cite{leach2022denoising,jagvaral2024unified}. A perturbation of an orientation $M$ is obtained by sampling a noise rotation $E\sim\mathcal{IG}_{\mathrm{SO}(3)}(I,\sigma^2)$ centred at the identity and composing the two rotations as $R=ME$. The resulting distribution is centred at $M$, with spread controlled by $\sigma$.
The rotational diffusion prior is given by $p(R^K)=\mathcal{IG}_{\mathrm{SO}(3)}(A,\sigma_K^2)$ and it depends on the observed state through the current state reference $A$.

For training, a clean rotation $R^0$ is corrupted directly at level $k$ by first computing the contracted centre $M^k$ and then adding a sampled noise rotation:
\begin{equation}
\label{eq:so3_forward}
\begin{aligned}
M^k &= A\,\mathrm{Exp}\!\left(\sqrt{\bar\alpha_k}\,\mathrm{Log}(A^\top R^0)\right),\\
E^k &\sim\mathcal{IG}_{\mathrm{SO}(3)}(I,\sigma_k^2),\qquad R^k=M^kE^k,
\end{aligned}
\end{equation}
where $\sigma_k^2=\sigma_{\max}^2(1-\bar\alpha_k)$. To interpret the first line, note that $R^0=A(A^\top R^0)$: $A^\top R^0$ is the rotation relative to $A$, and $\mathrm{Log}$ expresses it as an axis--angle vector. Multiplication by $\sqrt{\bar\alpha_k}$ reduces its angle. The exponential map converts the scaled vector back to a rotation matrix, which is composed with $A$ to obtain $M^k$. The second line adds noise by composing $M^k$ with $E^k$. As $\bar\alpha_k$ approaches zero, the centre approaches $A$ and the noise scale approaches $\sigma_{\max}$, yielding the prescribed prior in the limit.

For each rotation and frame, the network outputs $\hat\epsilon_\phi^{R,k}\in\mathbb R^3$, trained to predict the standardized angular noise $\epsilon^{R,k}=\mathrm{Log}((M^k)^\top R^k)/\sigma_k$. Scaling the prediction by $-\sigma_k$ gives the axis--angle correction that cancels the predicted noise. The exponential map converts this vector, through its skew-symmetric matrix in $\mathfrak{so}(3)$, into a rotation matrix. Applying this correction gives, before clipping,
\begin{equation}
\label{eq:so3_denoising_prediction}
\hat M^k=\hat R^k\,\mathrm{Exp}(-\sigma_k\hat\epsilon_\phi^{R,k}).
\end{equation}
This estimates the contracted centre $M^k$. To recover an estimate of the clean rotation relative to $A$, we divide $\mathrm{Log}(A^\top\hat M^k)$ by $\sqrt{\bar\alpha_k}$. We then combine this vector with the current noisy rotation in the same coordinates, using the DDPM coefficients:
\begin{equation}
\label{eq:so3_reverse}
\begin{aligned}
p_\phi(\hat R^{k-1}\mid\hat s^{D,k}_\tau)
&=\mathcal{IG}_{\mathrm{SO}(3)}(\mu^{R,k},\sigma_{\max}^2\tilde\beta_k),\\
\mu^{R,k}&=A\,\mathrm{Exp}\!\Big(
\frac{a_k}{\sqrt{\bar\alpha_k}}\mathrm{Log}(A^\top\hat M^k)\\
&\hspace{5.6em}+b_k\,\mathrm{Log}(A^\top\hat R^k)\Big),
\end{aligned}
\end{equation}
where $a_k=\sqrt{\bar\alpha_{k-1}}\beta_k/(1-\bar\alpha_k)$ and $b_k=\sqrt{\alpha_k}(1-\bar\alpha_{k-1})/(1-\bar\alpha_k)$. The resulting kernel is an approximation constructed in axis--angle coordinates. Sampling a reverse step amounts to multiplying its centre $\mu^{R,k}$ by a new identity-centred noise rotation. More details are provided in Appendix~\ref{app:pretrain_hparams}. At deployment, we use the predicted reverse means for both feature groups without adding noise at each step; generation remains stochastic through the initial prior sample.

\subsubsection{Generalist tracking policy}
\label{subsec:tracking_pretraining}

We train a single tracking policy on the feasible robot--object motions, optimizing~\eqref{eq:tracker_objective} with PPO in \texttt{mjlab}~\cite{zakka2026mjlablightweightframeworkgpuaccelerated}. The tracker operates at 50\,Hz using the residual joint targets in~\eqref{eq:tracker_interface}. Its reference observations include motion features at future offsets of $\{1,2,4,8\}$ frames. Appendix~\ref{app:tracking_training} specifies the observations and motion-sampling strategy; the rewards and termination conditions are defined in Appendices~\ref{app:tracking_rewards} and~\ref{app:tracking_terminations}.

\subsection{Removing Task Conditioning with SD:DPO}
\label{subsec:post_training}
\label{subsubsec:sdpo}

Although pretraining includes examples without reference conditioning through masking, removing the reference substantially reduces closed-loop survival (Section~\ref{subsec:res_cond}) since planning and tracking errors can lead the system into states poorly represented in the training data. Thus, at such states, the data provides limited guidance on how to continue the motion. To mitigate this problem we propose Self-Distillation Diffusion Policy Optimization (SD:DPO) to supply this supervision, where a student that receives only state history and the object goal learns to match the task-prior-conditioned teacher at states reached during environment rollout.

The student $G_\phi$ receives $c_t$ and the null reference token $\varnothing$, whereas the teacher $G_{\bar\phi}$ additionally receives the kinematic reference $s^{IK}_\tau$. Both are initialized from the pretrained planner. At each planning step, both generate a chunk. We choose either proposal with probability $1/2$ and execute its first $m$ frames with the frozen tracker. The resulting measured states update the history used to generate the next chunk.

At every planning step, we store the student's proposal, its starting context, and the corresponding kinematic reference in a buffer $\mathcal B$, regardless of which proposal was executed. Failed rollouts terminate and reset, but all previously collected proposals remain in the buffer, including those associated with failure. The tracker supplies the execution contexts used for learning while the teacher supplies the denoising targets.

Let $\phi_{\mathrm{old}}$ denote the student weights held fixed during collection. For each training update, we draw a stored proposal and corrupt it at a uniformly sampled diffusion level:
\begin{equation}
\label{eq:sdpo_sampling}
\begin{aligned}
\hat s^{D,0}_\tau&\sim G_{\phi_{\mathrm{old}}}(\cdot\mid c_t,\varnothing),
\qquad k\sim\mathcal U\{1,\ldots,K\},\\
\hat s^{D,k}_\tau&\sim q_k\bigl(\cdot\mid
\operatorname{sg}(\hat s^{D,0}_\tau),c_t\bigr).
\end{aligned}
\end{equation}
Here $q_k$ denotes direct corruption to level $k$ using the processes in Section~\ref{subsec:diffusion_pretraining}. Teacher and student receive the same noisy chunk, diffusion level, and stored context. The teacher also receives the kinematic reference, which provides guidance for predicting a continuation from that context. We transfer this guidance to the student by matching their Euclidean and rotational denoising predictions:
\begin{equation}
\label{eq:sdpo_practice}
\begin{aligned}
\mathcal L_{\mathrm{distill}}(\phi)
={}&\mathbb E_{\mathcal B,k,q}\Big[
\operatorname{MSE}\!\bigl(\hat{\mathbf v}^{k,\varnothing}_\phi,
\operatorname{sg}(\hat{\mathbf v}^{k,IK}_{\bar\phi})\bigr)\\
&\qquad+\operatorname{MSE}\!\bigl(\hat\epsilon^{R,k,\varnothing}_\phi,
\operatorname{sg}(\hat\epsilon^{R,k,IK}_{\bar\phi})\bigr)\Big].
\end{aligned}
\end{equation}
The superscripts $\varnothing$ and $IK$ indicate absent and present reference conditioning, respectively and each mean-squared error (MSE) averages the squared prediction differences over frames and features. The expectation is over buffered proposals, diffusion levels, and sampled noise. Importantly, gradients pass only through the student's predictions whereas teacher targets, buffered proposals, and noisy inputs are held fixed, so learning does not differentiate through collection, simulation, or noise sampling.

We train with~\eqref{eq:sdpo_practice} alongside the original data-based pretraining loss and an anchor loss. The anchor loss compares the student's reference-conditioned predictions with those of a frozen pretrained copy on the same noisy chunk. It penalizes deviations from the pretrained model, restricting changes to the reference-conditioned behavior. The teacher is updated through an exponential moving average of the student weights, while the pretrained copy remains fixed. Further details are provided in Appendix~\ref{app:sdpo_derivation}.

\subsection{Improving Task Completion with DA:PPO}
\label{subsubsec:da_ppo}

SD:DPO improves generation without a kinematic reference, but producing trackable motion does not by itself ensure that the object reaches its goal. We therefore fine-tune the distilled planner with DA:PPO. The frozen tracker executes its plans in simulation, and the resulting task rewards guide the planner's updates. The planner continues to receive only state history and the object goal, with the reference token set to $\varnothing$.

We instantiate DA-MDPs as explained in Section~\ref{subsec:damdp_preliminaries} with planner state $z_\tau=c_t$ and noisy action $\tilde a_\tau^k=\hat s^{D,k}_\tau$. The reward $r_\tau$ collected during execution of the first $m$ frames is assigned to the final denoising decision. The Bellman recursion then propagates this reward through the decisions that produced the chunk. If the episode continues, the planner generates a new chunk over $\tau'=t+m+1:t+m+n$ from the updated measured history.

We optimize~\eqref{eq:dappo_loss} to increase the execution return while retaining the log-ratio regularization. The tracker remains fixed throughout. Details on hyperparameters and rewards are provided in Appendices~\ref{app:posttraining_rewards} and~\ref{app:dappo_hparams}. IK-retargeted references can optionally be used to compute training rewards, but do not need to be used to condition the planner (see Sec.~\ref{subsec:res_reward}).



\begin{table*}[t]
      \centering
      \caption{Post-Training for Alignment on different pretraining corpora, evaluated in closed-loop using motions in the $Single$ dataset. Each generator is finetuned for 500 iterations on its own pretraining corpus, with 2048 parallel environments. The last two rows involve no learned planner: the same tracker is tasked to track the IK-retargeted motion or its SBTO solution.
  }
      \label{tab:damdp_corpus_ablation}
        \small
        \setlength{\tabcolsep}{8pt}
        \renewcommand{\arraystretch}{1.15}

        \begin{tabular}{llccccc}
        \toprule
        \textbf{Pretraining corpus} & \textbf{Reference source} & \textbf{Survival} & \textbf{Task success} & \textbf{MPKPE} & \textbf{Obj. pos.} & \textbf{Obj. ori.} \\
        & & \textbf{(\%) $\uparrow$} & \textbf{(\%) $\uparrow$} & \textbf{(cm) $\downarrow$} & \textbf{(cm) $\downarrow$} & \textbf{($^\circ$) $\downarrow$} \\
        \midrule
        \multirow{3}{*}{SBTO Aug + Symmetry} & Pretrained planner & 96.5 $\pm$ 1.5 & 51.9 $\pm$ 4.5 & 8.5 $\pm$ 0.6 & 12.6 $\pm$ 2.4 & 14.0 $\pm$ 1.3 \\
         & $+$ DPPO-style ($\mathcal{T} = 0$) & 99.1 $\pm$ 0.1 & 94.0 $\pm$ 3.5 & 4.9 $\pm$ 0.3 & 5.2 $\pm$ 2.1 & 6.3 $\pm$ 1.2 \\
         & $+$ DA:PPO & \textbf{100.0 $\pm$ 0.0} & \textbf{96.4 $\pm$ 3.4} & \textbf{4.3 $\pm$ 0.2} & \textbf{5.0 $\pm$ 2.0} & \textbf{5.8 $\pm$ 1.0} \\
        \midrule
        \multirow{2}{*}{SBTO Aug} & Pretrained planner & 92.3 $\pm$ 4.0 & 47.5 $\pm$ 6.3 & 8.6 $\pm$ 0.7 & 14.7 $\pm$ 3.9 & 14.9 $\pm$ 1.5 \\
         & $+$ DA:PPO & \textbf{100.0 $\pm$ 0.0} & \textbf{92.5 $\pm$ 4.7} & \textbf{4.1 $\pm$ 0.1} & \textbf{4.9 $\pm$ 1.9} & \textbf{5.8 $\pm$ 1.1} \\
        \midrule
        \multirow{2}{*}{Single + Symmetry} & Pretrained planner & 90.7 $\pm$ 4.9 & 49.1 $\pm$ 7.8 & 8.8 $\pm$ 0.8 & 18.7 $\pm$ 5.2 & 16.4 $\pm$ 1.8 \\
         & $+$ DA:PPO & \textbf{100.0 $\pm$ 0.0} & \textbf{89.4 $\pm$ 5.0} & \textbf{6.1 $\pm$ 0.3} & \textbf{7.2 $\pm$ 1.6} & \textbf{9.8 $\pm$ 1.3} \\
        \midrule
        \multirow{2}{*}{Single} & Pretrained planner & 90.0 $\pm$ 5.5 & 57.1 $\pm$ 7.8 & 9.0 $\pm$ 1.3 & 20.8 $\pm$ 6.7 & 18.3 $\pm$ 2.3 \\
         & $+$ DA:PPO & \textbf{93.3 $\pm$ 4.4} & \textbf{72.3 $\pm$ 7.5} & \textbf{8.8 $\pm$ 1.3} & \textbf{18.0 $\pm$ 6.8} & \textbf{15.0 $\pm$ 2.2} \\
         \midrule
         \multirow{2}{*}{None} & IK-retargeted motion & 93.5 $\pm$ 4.5 & 53.0 $\pm$ 8.3 & 8.2 $\pm$ 0.7 & 11.0 $\pm$ 3.4 & 31.4 $\pm$ 4.5 \\
         & SBTO solution & 100.0 $\pm$ 0.0 & 61.6 $\pm$ 8.0 & 6.6 $\pm$ 0.4 & 9.1 $\pm$ 2.4 & 20.3 $\pm$ 3.5 \\
        \bottomrule
        \end{tabular}
    \end{table*}

\section{Experiments and Results}
\label{sec:results}

We evaluate our pipeline on humanoid loco-manipulation tasks and organise the evaluations around the design choices. In our experiments, we report the \emph{Survival Rate} which we define as the fraction of rollouts where the robot neither falls nor drops the object. Additionally we define the \emph{Task success} as the fraction of survived rollouts whose realized final object pose lies
within $15$\,cm and $30^\circ$ of the goal position the motion generator was conditioned on. We also report \emph{MPKPE}, which is the mean per-keypoint position error of the executed robot motion against the generated motion from the diffusion model, and \emph{Obj.\ pos.}\ / \emph{Obj.\ ori.}\ the corresponding object position errors aggregated over the whole trajectory. Both metrics measure the trackability of the generated motions of the robot and object. All simulation results are reported as mean $\pm$ standard error of the mean (SEM) over distinct reference motions. Each motion is rolled out 100 times and averaged computing the error over different motions. 

\subsection{Choice of Pretraining Data Augmentations}
\label{subsec:res_corpus}

\begin{figure}[h]
    \centering
    \includegraphics[width=\linewidth]{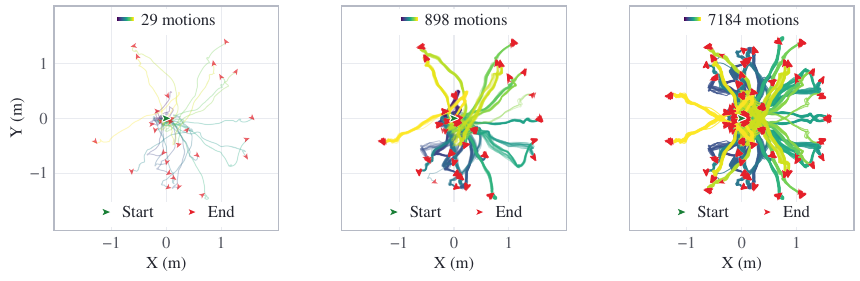}
    \caption{
    Planar robot trajectories aligned at the origin illustrating data augmentation for pick-carry-place task. Left: the original motions. Middle: dynamically feasible variations generated by SBTO from perturbed initial conditions. Right: the dataset after symmetry augmentation.
    }
    \label{fig:SBTO_aug}
\end{figure}

To identify important design choices during pretraining, we evaluate the individual and combined contributions of SBTO motion variations and symmetry augmentation in our pipeline (see Fig.~\ref{fig:SBTO_aug}). To test this, we pretrain diffusion models under the same training budget under four different settings: (i) using datasets containing a single motion per demonstration, (ii) additionally using symmetry augmentations, (iii) augmenting each motion with SBTO variations and (iv) combining each motion with both augmentations. SBTO variations are generated by repeating trajectory optimization with slightly perturbed initial conditions, producing dynamically feasible realizations of each demonstration with slight diversity. 
The resulting pretrained models are then evaluated in closed-loop simulation with the same pretrained tracking policy.

As shown in Table~\ref{tab:damdp_corpus_ablation} for the pretrained models, SBTO variations improve survival more than symmetry augmentation alone, while combining both yields the highest survival rate. These results indicate that diversity among dynamically feasible motions yields large individual benefit, with symmetries providing an additional improvement.
Table~\ref{tab:damdp_corpus_ablation} additionally shows that this advantage carries through post-training: using both augmentations, the survival rate reaches $100\%$, while substantially improving tracking performance and task success over not using any augmentations or just using symmetry augmentation. We also observe that, compared to DPPO~\cite{dppo2024}-style updates, DA:PPO~\cite{sanokowski2025damdp} yields a higher survival rate and better-tracking metrics, indicating that the maximum-entropy extension of DPPO to DA:PPO indeed improves performance. Furthermore, the results show that the pretrained planner achieves 51.9\% task success, while fine-tuning with DA:PPO improves this result to 96.4\%. The trackability of the IK-retargeted/SBTO motions are lower (see last row in Table~\ref{tab:damdp_corpus_ablation}) which yield a survival rate of $93.5\%$ and Task success of $53.0\%$.
Thus, the same pretrained tracking policy yields substantially higher task success when supplied with improved references from the finetuned planner.

\subsection{Importance of Task Conditioning During Pretraining}
\label{subsec:res_cond}


During pretraining, we find that using the IK-retargeted robot and object motion as conditioning for the diffusion planner, which provides a prior on how to solve the task, is important to obtain strong results after pretraining. To ablate this choice, we pretrain seven diffusion models on different task prior choices. The task priors range from using the full IK-retargeted motion down to not using retargeted conditioning at all. Each model is evaluated in closed-loop simulation using the same pretrained tracking policy.
  
\begin{figure}[h]
      \centering
      \includegraphics[width=\columnwidth]{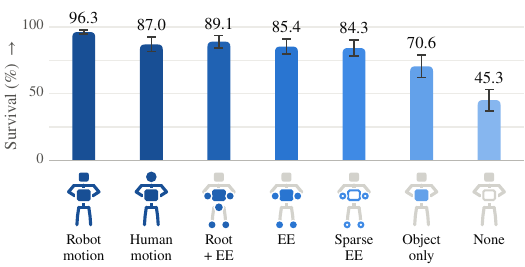} 
      \caption {Closed-loop survival rates of the pretrained generator under different conditioning priors. Each pictogram marks in blue what the prior specifies; hollow markers indicate sparse frames. 
  }
      \label{fig:cond-pickplace-anchor-success}
\end{figure}

As shown in Fig.~\ref{fig:cond-pickplace-anchor-success}, the strongest performance is obtained when using the IK-retargeted robot and object motion as task conditioning as derived from OmniRetarget dataset~\cite{yang2025omniretargetinteractionpreservingdatageneration}.
We also directly compare against using the corresponding human pose and object pose from the OMOMO~\cite{li2023omomo} dataset without any retargeting and observe a roughly $9\%$ drop in survival rate. 
Using sparse end-effector and object motion references (only every 5 frames), we do not observe a significant drop in performance, compared to using dense end-effector and object motion conditioning. 
Removing the remaining priors lowers the closed-loop survival rate further, to ${\sim}70\%$ with only the object motion, and to ${\sim}45\%$ with no task priors. These results suggest that, in our limited data regime, strong task priors are required during pretraining.

Later, in Section~\ref{subsec:res_sdpo}, we show that SD:DPO post-training removes the need for a strong task prior at deployment, enabling goal-conditioned motion generation without any performance sacrifice.

\subsection{SO(3) Diffusion Improves Performance}
\label{subsec:res_so3}


\begin{table}[h]
    \centering
    \small
    \caption{Comparison of diffusing the rotational components of the generated motion (root orientation, object orientation) on $\mathrm{SO}(3)$ vs.\ treating their $6$D coordinates as Euclidean. Evaluated closed-loop with the frozen tracker. 
    }
    \label{tab:so3-euclid-levels}
    \setlength{\tabcolsep}{6pt}
    \renewcommand{\arraystretch}{1.15}
    \begin{tabular}{@{}lccc@{}}
      \toprule
      \textbf{Rotation} & \textbf{Survival} & \textbf{Task success} & \textbf{Obj. ori.} \\
      \textbf{diffusion} & \textbf{(\%) $\uparrow$} & \textbf{(\%) $\uparrow$} & \textbf{($^\circ$) $\downarrow$} \\
      \midrule
      $\mathrm{SO}(3)$ & \textbf{96.5 $\pm$ 1.5} & \textbf{51.9 $\pm$ 4.5} & \textbf{14.0 $\pm$ 1.3} \\
      Euclidean & 85.9 $\pm$ 5.0 & 46.4 $\pm$ 5.7 & 19.6 $\pm$ 2.5 \\
      \bottomrule
      \end{tabular}
\end{table}

Standard diffusion models add Gaussian noise in Euclidean space and ignore the geometry of the manifold , whereas the root and object orientations in our motion representation lie on $\mathrm{SO}(3)$.
We therefore ablate on the choice of diffusion process for the rotational features.
Table~\ref{tab:so3-euclid-levels} compares two pretrained generators that differ only in this choice: one uses a VP-$\mathrm{SO}(3)$ diffusion process as described in Section~\ref{subsec:diffusion_pretraining}, the other one uses a VP-diffusion process in Euclidean space.
Our results show that diffusing on $\mathrm{SO}(3)$ raises the closed-loop survival rate from $85.9\%$ to $96.5\%$, while also improving across other tracking metrics.

\subsection{Motion-Level vs.\ Action-Level Diffusion}
\label{subsec:res_action}

Our pipeline separates motion generation from feedback control: a diffusion planner proposes whole-body and object trajectories, and an RL-trained tracking policy executes them. Alternatively, diffusion models can predict actions directly, either alone~\cite{truong2024pdp} or jointly with future states~\cite{Huang_2025,liao2025beyondmimicmotiontrackingversatile}.

We train these action-level baselines by behaviour cloning on state--action data collected by tracking the SBTO- and symmetry-augmented motions. Although this dataset includes perturbed states and corrective actions, the learned feedback behaviour depends on the coverage of a fixed dataset. Our planner also learns from a limited motion corpus, but its tracker learns via RL in simulation, where it can learn to respond to disturbances and errors encountered during its own execution. This separation allows us to learn motion structure from the available data while optimizing feedback control through interaction.

The learning objectives also favour different behaviours. From a maximum-likelihood perspective, generative learning corresponds to forward-KL fitting, which encourages coverage of the data distribution. Under imperfect approximation, a model may cover multiple action modes while also assigning probability to low-density regions between them~\cite{minka2005divergence}. For an action-level model, such samples become joint commands and may disrupt balance or contact.

Maximum-entropy RL instead admits a reverse-KL formulation with a reward-weighted target distribution~\cite{levine2018reinforcement}. This objective discourages probability mass on low-return behaviours and can favour concentrating on a subset of successful behaviours. Although this mode-seeking tendency can hinder exploration of alternative behaviours, it favours reliable execution when failures receive low reward. This motivates combining a generative motion planner with an RL-trained tracker: the planner represents motion diversity, while the tracker learns to select actions according to their consequences in the environment.

We therefore compare motion-level and action-level diffusion using the same underlying motion corpus and evaluation tasks to assess whether separating planning from feedback control improves closed-loop survival and task success.

\begin{table}[h]
    \centering
    \caption{Diffusion \emph{motion} generator with tracking policy vs.\ diffusion \emph{action} generators executing joint PD targets directly for loco-manipulation tasks.
    The state-action diffusion baseline is an unconditioned prior that is steered at test time.
    }
    \label{tab:motion_vs_action_generator}
    \small
    \setlength{\tabcolsep}{4pt}
    \renewcommand{\arraystretch}{1.15}
    \begin{tabularx}{\columnwidth}{@{}>{\raggedright\arraybackslash}X cc@{}}
        \toprule
        \textbf{Model} & \textbf{Survival (\%) $\uparrow$} & \textbf{Task success (\%) $\uparrow$} \\
        \midrule
        State Diffusion + Tracker
        & \textbf{96.5 $\pm$ 1.5}
        & \textbf{51.9 $\pm$ 4.5} \\

        Action Diffusion~\cite{truong2024pdp}
        & 87.5 $\pm$ 3.7
        & 45.3 $\pm$ 4.8 \\

        \mbox{State-Action Diffusion~\cite{liao2025beyondmimicmotiontrackingversatile}}
                & 58.2 $\pm$ 7.1
                & 8.2 $\pm$ 3.6 \\

        \bottomrule
    \end{tabularx}
\end{table}

We consider two baselines that output robot joint PD targets directly, with no tracker in the loop: an action-only diffusion policy as in PDP~\cite{truong2024pdp}, and a state--action diffusion model as in BeyondMimic~\cite {liao2025beyondmimicmotiontrackingversatile} for humanoid control.
Both are trained via behaviour cloning on rollouts collected by our pretrained generalist tracking policy over the SBTO- and symmetry-augmented data corpus, and both replan at every control step ($50$\,Hz), executing only the first action.
In this comparison all approaches receive the same task information which consist of states, the goal object pose and the IK-retargeted robot and object motion.
We also compare to~\cite{liao2025beyondmimicmotiontrackingversatile}, where a state-action diffusion model is instead an unconditioned prior that is steered at inference. We adopt this by steering the diffusion model towards the IK-retargeted motion with keyframe in-betweening.
See Appendix~\ref{app:action_baselines} for more details.

The results are presented in Table~\ref{tab:motion_vs_action_generator}, where the PDP-style policy is the strongest action-diffusion baseline, with a survival rate of $87.5\%$ compared to $96.5\%$ for our pretrained diffusion planner combined with the tracker. Our approach also achieves significantly better task success.
State-action diffusion is overall the weakest baseline: the steered prior reaches a survival rate of only $58.2\%$ and a task success of $8.2\%$.
On both metrics, our hierarchical approach, a state-only diffusion model for motion generation paired with a high-frequency tracking policy, performs best, even before fine-tuning.

Furthermore, inference of the diffusion model on our hardware setup, with the tracker running in a parallel process, runs at ~12.5 Hz. Although action/state-action diffusion models can plan actions over a longer horizon, the generated actions operate open-loop until the next plan arrives, making it extremely challenging for legged robots to stabilise at such low frequencies, compared to having a separate high-frequency low-level tracking policy as we do.

\subsection{RL fine-tuning reward design}
\label{subsec:res_reward}



\begin{table}[h]
      \centering
      \caption{DA:PPO finetuning with the IK-motion tracking reward $r_\text{kin}$ vs.\ a goal-driven reward $r_\text{goal}$.
      }
      \label{tab:damdp_reward}
      \small
      \setlength{\tabcolsep}{6pt}
      \renewcommand{\arraystretch}{1.15}
      
      \begin{tabular}{@{}lccc@{}}
      \toprule
      & & \multicolumn{2}{c}{\textbf{$+$ DA:PPO}} \\
      \cmidrule(l){3-4}
      \textbf{Metric} & \textbf{Pretrained} & \textbf{$r_\text{kin}$} & \textbf{$r_\text{goal}$} \\
      \midrule
      Survival (\%) $\uparrow$ & 96.5 $\pm$ 1.5 & \textbf{100.0 $\pm$ 0.0} & \textbf{100.0 $\pm$ 0.0} \\
      Task success (\%) $\uparrow$ & 51.9 $\pm$ 4.5 & \textbf{96.4 $\pm$ 3.4} & 92.9 $\pm$ 4.7 \\
      MPKPE (cm) $\downarrow$ & 8.5 $\pm$ 0.6 & \textbf{4.3 $\pm$ 0.2} & 8.0 $\pm$ 0.4 \\
      Obj. pos. (cm) $\downarrow$ & 12.6 $\pm$ 2.4 & \textbf{5.0 $\pm$ 2.0} & 7.2 $\pm$ 2.0 \\
      Obj. ori. ($^\circ$) $\downarrow$ & 14.0 $\pm$ 1.3 & \textbf{5.8 $\pm$ 1.0} & 7.5 $\pm$ 1.3 \\
      \bottomrule
  \end{tabular}
 
  \end{table}

Next, we investigate how the design of the rewards for diffusion RL finetuning with DA:PPO affects the performance of the overall pipeline.
A natural choice is a dense reward that encourages resulting rollouts to stay close to the IK-retargeted motion. Such a reward is similar to the whole-body and object-tracking rewards used to train the tracking policy, but it can only be computed where a retargeted demonstration is available.
An alternative approach is a simpler goal-driven reward, which does not need a corresponding kinematic trajectory.

In Table~\ref{tab:damdp_reward}, we compare finetuning with DA:PPO as described in Section~\ref{subsubsec:da_ppo}, starting from the same pretrained diffusion motion generator using the two different rewards described above: $r_\text{kin}$, which rewards staying close to the dense IK-retargeted motion at every step, and $r_\text{goal}$, which rewards approaching the goal combined with a survival bonus (see Appendix~\ref{app:posttraining_rewards} for details). 
Both reward sets improve survival and task success relative to the pretrained model. $r_\text{kin}$ furthermore reduces all reported tracking errors with respect to generated motions over using $r_\text{goal}$. $r_\text{goal}$ achieves 92.9\% task success, compared with 96.4\% for using $r_\text{kin}$, but increases MPKPE and other tracking errors. Although, we stick to using $r_{\text{kin}}$ for our experiments, $r_\text{goal}$ could also be used for finetuning towards unseen goals using a curriculum and where demonstrations are unavailable.


\subsection{Effect of Maximum-Entropy Fine-Tuning}
\label{subsec:res_objective}

In Table~\ref{tab:damdp_corpus_ablation}, we also compare the finetuning capabilities of DA:PPO against a DPPO-style update, which ablates on using DA:PPO without temperature (see Section~\ref{subsubsec:da_ppo}). The original DPPO~\cite{dppo2024} uses additional tricks, such as injecting more diffusion noise for exploration, which we removed because they led to unstable and untrackable behaviour. While the DPPO-style update improves overall metrics compared to the pretrained model, we find that using the temperature in DA:PPO updates improves the survival rate to $100\%$ and significantly improves all other metrics when finetuned under identical settings and hyperparameters. These improvements can be attributed to the maximum entropy RL formulation of DA:PPO which encourages exploration while the DPPO-style objective suffers under severe mode collapse~\cite{sanokowski2025damdp}.

\subsection{Removing the need for strong task conditioning}
\label{subsec:res_sdpo}


\begin{table}[h]
    \centering
    \caption{Diffusion motion generators post-trained using SD:DPO on different tasks. 
    \emph{Self-Teacher} and \emph{Student} are the same pretrained generator, evaluated with the IK-retargeted task prior and with the null token $\varnothing$ in its place, respectively. \emph{No-prior} is a separate motion generator described in Section~\ref{subsec:res_cond} that doesn't receive any task prior upon pretraining. 
    }
    \label{tab:sdpo_tasks}
    \footnotesize
    \setlength{\tabcolsep}{4pt}
    \renewcommand{\arraystretch}{1.15}

















    
     \begin{tabular}{@{}llcc@{}}
      \toprule
      \textbf{Task} & \textbf{Generator} & \textbf{Survival} & \textbf{Task success} \\
      & & \textbf{(\%) $\uparrow$} & \textbf{(\%) $\uparrow$} \\
      \midrule

      \multirow{8}{*}{Pick-place}
      & No-prior (pretrained)
      & 46.9 $\pm$ 8.3 & 32.2 $\pm$ 6.8 \\

      & No-prior $+$ DA:PPO ($r_\text{kin}$)
      & 74.1 $\pm$ 8.0 & 67.0 $\pm$ 8.6 \\
      \cmidrule(l){2-4}
      & \emph{Self-Teacher (pretrained)}
      & 96.5 $\pm$ 1.5 & 51.9 $\pm$ 4.5 \\

      & Student (pretrained)
      & 57.7 $\pm$ 9.2 & 45.4 $\pm$ 9.0 \\

      & Student $+$ DA:PPO
      & 92.1 $\pm$ 4.8 & 92.1 $\pm$ 4.8 \\

      & Student $+$ SD:DPO
      & \textbf{100.0 $\pm$ 0.0} & 55.1 $\pm$ 6.0 \\


      & Student $+$ SD:DPO $+$ DA:PPO
      & 99.5 $\pm$ 0.5 & \textbf{98.8 $\pm$ 1.2} \\

      \midrule

      \multirow{4}{*}{Kick}
      & \emph{Self-Teacher (pretrained)}
      & 99.9 $\pm$ 0.1 & 34.4 $\pm$ 7.2 \\

      & Student (pretrained)
      & \textbf{100.0 $\pm$ 0.0} & 1.2 $\pm$ 0.7 \\

      & Student $+$ SD:DPO
      & \textbf{100.0 $\pm$ 0.0} & 31.9 $\pm$ 7.1 \\

      & Student $+$ SD:DPO $+$ DA:PPO
      & \textbf{100.0 $\pm$ 0.0} & \textbf{68.1 $\pm$ 8.3} \\

      \midrule

      \multirow{4}{*}{Push/Pull}
      & \emph{Self-Teacher (pretrained)}
      & \textbf{100.0 $\pm$ 0.0} & 23.3 $\pm$ 4.9 \\

      & Student (pretrained)
      & 98.7 $\pm$ 1.3 & 12.8 $\pm$ 3.9 \\

      & Student $+$ SD:DPO
      & \textbf{100.0 $\pm$ 0.0} & 21.7 $\pm$ 4.8 \\

      & Student $+$ SD:DPO $+$ DA:PPO
      & \textbf{100.0 $\pm$ 0.0} & \textbf{52.7 $\pm$ 6.5} \\

      \bottomrule
  \end{tabular}
\end{table}

We have demonstrated in Sec.~\ref{subsec:res_cond} that pretraining without any task prior yields significantly lower task performance compared to using the task priors. However, this might limit the applicability of our method as it requires IK retargeted references from human motions, which is not always available. 
To mitigate this we show in Table \ref{tab:sdpo_tasks} that SD:DPO (see Sec.~\ref{subsubsec:sdpo}) enables us to achieve high survival rates and high task success, without relying on the IK-retargeted task prior. 

Table \ref{tab:sdpo_tasks} separates the effects of self-distillation and subsequent finetuning via diffusion-RL. For pick-place, SD:DPO raises survival of the \emph{Student} model from 57\% to 100\%. While finetuning the pretrained \emph{Student} via DA:PPO increases the task success to 92\%, further finetuning of the self-distilled \emph{Student} increases the task success to $\sim$99\%. We also show that the no-prior model cannot improve significantly via finetuning compared to finetuning the self-distilled student, highlighting the importance of using the task conditioning during pretraining. 

Additionally, we reproduce our experiments on an alternate set of challenging motions that involve kicking the object or pushing/pulling towards a target goal pose. For each of these, we pretrain a diffusion motion generator only on the corresponding motions, following Section~\ref{subsec:diffusion_pretraining}, then post-train via self-distillation (SD:DPO) to eliminate conditioning on task priors, and finally post-train for fine-tuning to further improve task performance. We observe trends similar to those when using only pick-carry-place task motions, confirming that our method is general and can also be applied to very different tasks.

\begin{figure}[t]
      \centering
      \includegraphics[width=\columnwidth]{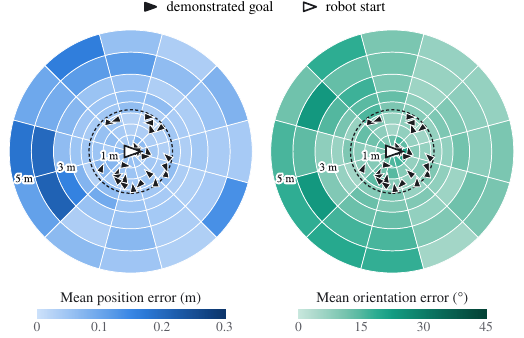}
      \caption{Mean final position error (left) and orientation error (right) of the object for Student $+$ SD:DPO $+$ DA:PPO, with goals placed around the robot's start. The robot starts at the center facing right; black arrowheads are the $29$ demonstrated goals, pointing along the goal yaw, and the dashed circle marks the farthest of them ($1.9$\,m). Each goal pose is evaluated using $100$ rollouts, with the objects's initial pose randomised up to $5$\,cm and $12^\circ$.}
      \label{fig:goal_reach_map}
  \end{figure}

By removing dependence on task conditioning, we can also plan motions using new goal poses that are unseen in our dataset. To measure this generalisation systematically, in Figure~\ref{fig:goal_reach_map}, we place goals for the object on a polar grid around the robot's start at distances ranging from $0.5$ to $5$\,m, and each position has $0^\circ/90^\circ/180^\circ/270^\circ$ rotations of the goal. The farthest goal in our demonstration set is at $1.9$ from the origin (marked with dashed circle). We find that we can achieve these goals with $\sim$100\% survival rates and a high task success rate of $88\%$ on this grid test for generalisation. Further plots in Appendix~\ref{app:goal_reach_maps} show the same evaluation for the pretrained \emph{Student} and \emph{Student $+$ SD:DPO}, along with example robot trajectories for individual new goals.

\subsection{Planner Finetuning beats Tracker Finetuning}
\label{subsec:res_tracker_ft}


Instead of finetuning the diffusion planner with DA:PPO to close the plan--execution gap, it is also possible to finetune the tracking policy towards better execution of the diffusion plan, as done in \cite{zhang2026learning}. 
As we show in this section, however, we observe that diffusion planner finetuning leads to better task success and better survival rates.
To ablate on this we finetune the pretrained tracking policy while keeping the motion generator frozen (see Appendix~\ref{app:kl_lora}). We fine-tune the tracking policy with and without the goal term, which yield similar results. We also ablate only finetuning the tracker on the pretrained diffusion motion generator against finetuning the tracker using the finetuned diffusion motion generator.

\begin{table}[h]
    \centering
    \caption{Comparison of fine-tuning the motion generator against fine-tuning the tracker. In the top block we use the pretrained diffusion model, and in the bottom block the diffusion model post-trained with DA:PPO. Tracker fine-tuning updates all parameters of the tracking policy with PPO for 1000 learning iterations, using the reward for tracking the generated reference ($r_\text{gen}$), with or without an additional goal term ($r_\text{goal}$); see Appendix~\ref{app:posttraining_rewards}.
    }
    \label{tab:tracker_ft}
    \small
    \setlength{\tabcolsep}{3pt}
    \renewcommand{\arraystretch}{1.15}
    \begin{tabular*}{\columnwidth}{@{\extracolsep{\fill}}llcc@{}}
        \toprule
        \textbf{Motion}    & \textbf{Tracker} & \textbf{Survival}         & \textbf{Task success} \\
        \textbf{generator} &                  & \textbf{(\%) $\uparrow$} & \textbf{(\%) $\uparrow$} \\
        \midrule
        \multirow{3}{*}{Pretrained}
            & Pretrained                              & 96.8 $\pm$ 2.2          & 51.9 $\pm$ 4.5 \\
            & PPO ($r_\text{gen}$)                    & 99.0 $\pm$ 1.0          & 87.9 $\pm$ 4.9 \\
            & PPO ($r_\text{gen}{+}r_\text{goal}$)    & \textbf{99.5 $\pm$ 0.3} & \textbf{89.3 $\pm$ 5.2} \\
        \midrule
        $+$\,DA:PPO ($r_\text{kin}$)
            & Pretrained                              & \textbf{100.0 $\pm$ 0.0} & \textbf{96.5 $\pm$ 1.5} \\
        \bottomrule
    \end{tabular*}
\end{table}

As shown in Table~\ref{tab:tracker_ft}, with the pretrained tracker fixed, planner fine-tuning increases task success from 51\% to 96\%. Fine-tuning the tracker with the pretrained planner reaches at most 89\%. These results show that finetuning the diffusion planner alone already recovers large improvement in task success, with the tracker left unchanged.

\subsection{Real-World Deployment}
\label{subsec:res_real}

\begin{table}[h]
    \centering
    \caption{Real-robot execution on the Unitree G1, for ten executions of a single goal (\emph{repeated}) and one execution of distinct goals (\emph{distinct}). Success is the number of successful over attempted executions; the remaining values are mean $\pm$ sample standard deviation over executions.}
    \label{tab:real_robot_motion_groups}
    \small
    \setlength{\tabcolsep}{4pt}
    \renewcommand{\arraystretch}{1.15}
    \begin{tabular}{@{}lcccc@{}}
        \toprule
        \textbf{Motions} & \textbf{Success} & \textbf{MPKPE} & \textbf{Final obj. pos.} & \textbf{Final obj. ori.} \\
        & \textbf{$\uparrow$} & \textbf{(cm) $\downarrow$} & \textbf{(cm) $\downarrow$} & \textbf{($^\circ$) $\downarrow$} \\
        \midrule
        Repeated & 10/10 & 6.5 $\pm$ 0.7 & 1.8 $\pm$ 0.8 & 3.4 $\pm$ 1.1 \\
        Distinct & 5/5 & 6.8 $\pm$ 1.1 & 5.7 $\pm$ 5.0 & 12.8 $\pm$ 12.5 \\
        \bottomrule
    \end{tabular}
\end{table}

We evaluate our diffusion motion generator and tracking policy on a $29$-DoF Unitree G1 humanoid performing pick-carry-place tasks as shown in Fig.~\ref{fig:real_robot}. 
The motion generator runs on the GPU in a separate process, while the
tracking executes on the CPU through ONNX Runtime. 
Joint encoders and the onboard IMU provide proprioceptive states, whereas a motion-capture system tracks the robot base and object, providing robot position
and object pose for closed-loop execution.

The tracking policy runs at $50$\,Hz and outputs residual joint positions that are added to the generated reference to form joint-level PD targets, which are tracked by a PD controller running at $200$\,Hz.
The generator uses $10$ denoising steps to predict a chunk of $64$ robot--object motion, corresponding to a $1.28$\,s horizon.
The tracker continues executing the current chunk from the generator during inference, until a new chunk arrives from the generator process. 
In our setup, the motion generator can run at a maximum frequency of $\sim12.5$\,Hz, accounting for communication latency and diffusion sampling.

Table~\ref{tab:real_robot_motion_groups} reports two groups of experiments: (a) ten repeated executions of the same final goal pose, and (b) one execution for distinct final goal poses. 
Every attempted execution succeeds: ten out of ten times for the repeated motion, and five out of five across the distinct motions.

\section{Conclusion and Future Work}
We presented DiPLO and established a key finding for limited-data humanoid loco-manipulation: strong task conditioning can be crucial for effective pretraining, while post-training can remove this dependency for deployment. Starting from only 29 retargeted pick-carry-place demonstrations, dynamically feasible data augmentation and strong kinematic conditioning enable the diffusion planner to learn trackable whole-body and object motions. Removing this conditioning directly causes a substantial performance drop, while training from scratch without it cannot recover comparable performance through downstream fine-tuning alone.
We address this dependence with SD:DPO, which distills a reference-conditioned self-teacher into a planner conditioned only on state history and the object goal. This recovers goal-conditioned motion generation and enables the planner to operate on goals unseen in pretraining demonstrations. Complementary to this, DA:PPO further optimises the planner through closed-loop interaction with the frozen tracker, aligning generated motions with what the robot can successfully execute. Together, SD:DPO and DA:PPO achieve approximately 99\% survival and 99\% goal success during pick-carry-place rollouts, while the same training recipe transfers to substantially different skills, including kicking and pushing/pulling. The resulting planner-tracker also transfers directly to the Unitree G1 without any real-world fine-tuning.
These results suggest a broader principle for robot learning: privileged or expensive supervision may be most valuable as a scaffold for pretraining rather than as a permanent deployment requirement. Future work will investigate whether this principle extends to a single model trained jointly across multiple loco-manipulation skills, more diverse objects and environments, and onboard perception, thereby removing the current reliance on external motion capture.

\bibliographystyle{IEEEtran}
\bibliography{ref}

\clearpage




\title{Supplementary Material for\\
  DiPLO: Self-distilled \underline{Di}ffusion \underline{P}lanning for \underline{Lo}co-manipulation}
\author{Anonymous Authors}
\maketitle

\appendix

\subsection{Denoiser Network Architecture}
\label{app:architecture}

\begin{figure*}[t]
    \centering
    \includegraphics[width=\textwidth]{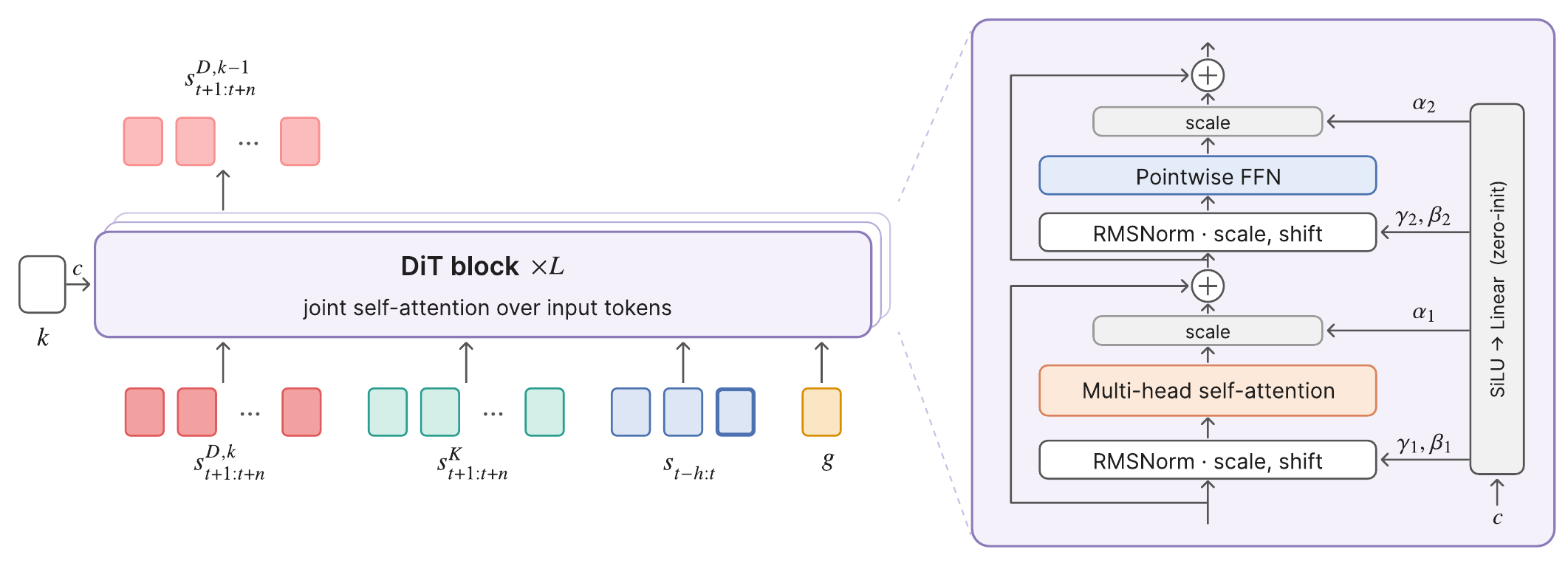}

    \caption{Architecture of the denoiser network of the diffusion-based motion generator $G_\phi$, based on $L$-layers of Diffusion Transformer (DiT) blocks. At each denoising step $k$, along with the noised target tokens $s^{D,k}_{\tau}$ (the motion chunk over the frames $\tau = t{+}1{:}t{+}n$ at denoising step $k$) and the denoising step $k$ via adaptive layer norm, the network is also conditioned on the goal token, the current state and history tokens, and the IK-retargeted motion token which serves as the prior to solve the task. The output is the denoised target $s^{D,k-1}_{\tau}$, and $G_\phi$ iterates the network from $k = K$ down to $k = 1$, whose output $s^{D,0}_{\tau}$ is the fully denoised motion.}
    \label{fig:denoiser}
\end{figure*}

Figure~\ref{fig:denoiser} shows the denoiser network of the diffusion motion generator (Section~\ref{subsec:diffusion_pretraining}).
Each input is tokenised by embedding to the same dimension using an MLP and accompanied with positional encoding for time series data. The denoising step $k$ enters the network through adaptive layer normalization as in~\cite{peebles2023dit}.

\subsection{ {Diffusion Motion Generator Pretraining}}
\label{app:pretrain_hparams}

{ 
\textit{Euclidean training loss.} Each training example consists of a dataset motion chunk with its conditioning, a uniformly sampled diffusion level $k\in\{1,\ldots,K\}$, and forward noise drawn from $q$. The denoiser predicts the $\mathbf v$-target defined in Section~\ref{subsubsec:diffusion_processes}. We penalize each scalar prediction error $r$ using smooth-L1 with a unit transition threshold:
\begin{equation}
\label{eq:smooth_l1}
\ell_{\mathrm{SL1}}(r)=
\begin{cases}
\tfrac12r^2,&|r|\leq1,\\
|r|-\tfrac12,&|r|>1.
\end{cases}
\end{equation}
For small errors, the quadratic penalty gives a gradient proportional to the error. For larger errors, the penalty grows linearly and its derivative has magnitude one, limiting the influence of large errors on the scalar loss gradient. Both the penalty and its derivative are continuous at the transition. The threshold is measured in $\mathbf v$-prediction units, computed from normalized features.

Let $\mathcal C_{\mathrm E}$ denote the Euclidean feature groups and $N_c$ the number of scalar entries across all frames and features in group $c$. For errors $r_{c,j}=\hat v_{\phi,c,j}^k-v_{c,j}^k$, we average within each group:
\[
\operatorname{SmoothL1}(\hat{\mathbf v}_{\phi,c}^k,\mathbf v_c^k)
=\frac{1}{N_c}\sum_{j=1}^{N_c}\ell_{\mathrm{SL1}}(r_{c,j}).
\]
The resulting Euclidean objective is
\begin{equation}
\label{eq:euc_loss}
\mathcal L_{\mathrm E}(\phi)
=\mathbb E_{s^D_{\tau},k,q}\!\left[
\omega_k\sum_{c\in\mathcal C_{\mathrm E}}\lambda_c\,
\operatorname{SmoothL1}(\hat{\mathbf v}_{\phi,c}^k,\mathbf v_c^k)
\right].
\end{equation}
The expectation averages over dataset chunks, diffusion levels, and forward noise. Averaging within groups makes their relative importance depend on the weights $\lambda_c$ in Table~\ref{tab:pretrain_hparams}, rather than their number of entries.

The diffusion-level weight is $\omega_k=1$ without noise-level reweighting. When Min-SNR weighting~\cite{hang2023minsnr} is enabled, we use $\omega_k=\min(\mathrm{SNR}_k,\gamma_{\mathrm{SNR}})/(1+\mathrm{SNR}_k)$, where $\mathrm{SNR}_k=\bar\alpha_k/(1-\bar\alpha_k)$ is the signal-to-noise ratio and $\gamma_{\mathrm{SNR}}$ is its clipping threshold. These weights adjust the contribution of different diffusion levels.

Smooth-L1 changes the regression objective: its optimal prediction need not equal the conditional mean of $\mathbf v^k$. The exact score-recovery argument for squared-error regression therefore does not apply to this loss.

\textit{Rotational training and sampling.} To sample the forward noise $E^k$ in~\eqref{eq:so3_forward}, we draw an axis uniformly on the unit sphere and an angle from a numerically tabulated $\mathcal{IG}_{\mathrm{SO}(3)}$ angle distribution at scale $\sigma_k$. Multiplying the axis by the angle gives a vector whose exponential is $E^k$.

}
Table~\ref{tab:pretrain_hparams} lists the full configuration used to pretrain the diffusion motion generator (Section~\ref{subsec:diffusion_pretraining}).

\begin{table}[!h]
    \centering
    \caption{Pretraining configuration of the diffusion motion generator.}
    \label{tab:pretrain_hparams}
    \small
    \setlength{\tabcolsep}{4pt}
    \renewcommand{\arraystretch}{1.1}
    \begin{tabularx}{\columnwidth}{@{}Xl@{}}
        \toprule
        \textbf{Parameter} & \textbf{Value} \\
        \midrule
        \multicolumn{2}{l}{\textit{Representation}} \\
        \midrule
        Data rate & $50$\,Hz \\
        History / prediction horizon $n_h$ / $n$ & $2$ / $64$ frames \\
        Per-frame feature dimension & $48$ \\
        Goal dimension $g_t$ & $10$ \\
        Feature normalization & per-channel mean/std \\
        \midrule
        \multicolumn{2}{l}{\textit{Denoiser}} \\
        \midrule
        Layers / heads & $4$ / $4$ \\
        Width / feed-forward width & $256$ / $1024$ \\
        Dropout & $0.1$ \\
        Conditioning & tokens (no cross-attention) \\
        \quad history / current frames & $2$ / $1$ \\
        \quad kinematic reference frames & $64$ \\
        \quad goal token & $1$ \\
        Sequence length (conditioning + target) & $68 + 64$ \\
        Query/key normalization & RMS, per head \\
        Block modulation & AdaLN-Zero (timestep only) \\
        Timestep embedding & sinusoidal \\
        Trainable parameters & $5.1$M \\
        \midrule
        \multicolumn{2}{l}{\textit{Diffusion process}} \\
        \midrule
        Denoising steps $K$ (train and inference) & $10$ \\
        Euclidean schedule & cosine VP \\
        Prediction target (Euclidean) & $\mathbf{v}$ \\
        Reverse sampler & DDPM (kernel mean at deployment) \\
        Predicted clean-sample clipping & $\pm 5\sigma$ \\
        Rotation forward kernel & $\mathcal{IG}_{\mathrm{SO}(3)}$, VP \\
        Prediction target (rotations) & tangent noise \\
        $\sigma_{\max}$ root residual (anchor: identity) & $0.4$\,rad \\
        $\sigma_{\max}$ object rotation (anchor: current) & $0.2$\,rad \\
        \midrule
        \multicolumn{2}{l}{\textit{Loss weights $\lambda_c$}} \\
        \midrule
        Joint positions & $20$ \\
        Root displacement & $10$ \\
        Root height / object position / orientation & $5$ \\
        Root residual orientation & $3$ \\
        Heading change & $2$ \\
        \midrule
        \multicolumn{2}{l}{\textit{Optimization}} \\
        \midrule
        Optimizer & AdamW \\
        Learning rate / weight decay & $10^{-4}$ / $0.05$ \\
        LR schedule & cosine, $1000$ warmup, $\to 1\%$ \\
        Batch size (windows) & $1024$ \\
        Gradient steps & $150$k \\
        Gradient-norm clip & $0.5$ \\
        EMA decay & $0.9995$ \\
        \bottomrule
    \end{tabularx}
\end{table}

\subsection{Goal Representation}
\label{app:finetune_dr}

$g_t = \left[\, \hat{d}_t,\; \varphi(\Delta R_t),\; \min\!\left(1.0, \lVert \Delta p_t \rVert\right) \right] \in \mathbb{R}^{10}$

Here $\Delta p_t = p^o_T - p^o_t$ is the displacement of the object towards its final pose, $\hat{d}_t \in \mathbb{R}^{3}$ is its unit direction expressed in the robot base frame, $\lVert \Delta p_t \rVert$ is clipped at $1$\,m, and $\varphi(\Delta R_t) \in \mathbb{R}^{6}$ is the 6D representation of the relative object rotation $\Delta R_t = (R^o_t)^{-1} R^o_T$, i.e.\ the rotation that still has to be applied to the object expressed in its own current frame.
Expressing the goal as a displacement from the \emph{current} object pose rather than as an absolute target keeps it on the same relative footing as the motion representation of Section~\ref{subsec:diffusion_pretraining}, and makes it a residual that shrinks to zero as the task is completed.

\begin{table}[!h]
    \centering
    \caption{Domain randomization ranges used during finetuning of the diffusion motion generator. Values denote symmetric half-widths of uniform perturbations.}
    \label{tab:dr1}
    \footnotesize
    \setlength{\tabcolsep}{3pt}
    \renewcommand{\arraystretch}{1.1}
    \begin{tabular}{@{}llc@{}}
        \toprule
        \textbf{Stage} & \textbf{Quantity} & \textbf{Range} \\
        \midrule
        \multirow{3}{*}{Reset} & Object position $(x,y,z)$ & $(\pm0.05,\pm0.05,\pm0.0125)$ m \\
        & Object orientation (r,p,y) & $(\pm0.05,\pm0.05,\pm0.2)$ rad \\
        & Robot root pose $(p,r)$ & $(\pm0.05,\pm0.1)$ m, rad \\
        & Robot root height / yaw & $\pm0.01$ m / $\pm0.2$ rad \\
        & Robot joint positions & $\pm0.1$ rad \\
        \midrule
        \multirow{3}{*}{Push} & Robot linear / angular velocity & $\pm0.3$ m/s / $\pm0.45$ rad/s \\
        & Object linear / angular velocity & $\pm0.35$ m/s / $\pm0.9$ rad/s \\
        & Push interval & $\mathcal{U}(2,5)$ s \\
        \bottomrule
    \end{tabular}
\end{table}

\subsection{Tracking Controller Training Details}
\label{app:tracking_training}

This appendix gives the observation layout and the reset strategy used to pretrain the generalist tracking controller (Section~\ref{subsec:tracking_pretraining}).

\textit{Observations.} The reference observations of the actor consist of reference joint states, body poses, and object-tracking features at future trajectory steps $\{1,2,4,8\}$ frames ($20$--$160$\,ms). We use an asymmetric actor--critic: the critic additionally receives object velocities, reference contact indicators, and a wider temporal window of reference motion to track.

\textit{Reset sampling.} Training resets select both a reference motion and a starting phase, then initialize the robot and object from that reference state with perturbations.
These two sampling decisions use separate mechanisms.
For the starting phase, we use adaptive reference-state initialization as in~\cite{liao2025beyondmimicmotiontrackingversatile} during the first $25{,}000$ learning iterations and uniform phase sampling thereafter.
For motion selection, we retain evaluation-based reweighting throughout training, including after the phase-sampling switch. Every $1000$ learning iterations, each GPU evaluates the motions in its local shard from the first frame without any domain randomization, distributing its environments across those motions.

A motion is classified as failed if \emph{any} of its evaluation triggers a non-timeout termination which entails robot and object terminations based on tracking as described in Appendix~\ref{app:tracking_terminations}.
Let $\mathcal{S}_g$ be the motion shard on GPU $g$, $N_g=|\mathcal{S}_g|$, and $F_i^{(m)}$ indicate failure at evaluation $m$.
Starting from unit weights, the update is
\begin{equation}
\label{eq:tracker_motion_weights}
 w_i^{(m+1)} =
 \begin{cases}
  1, & F_i^{(m)}=1,\\[2pt]
  \max\!\left(\gamma_s\,w_i^{(m)},\,N_g^{-1}\right), & F_i^{(m)}=0,
 \end{cases}
\end{equation}
and subsequent resets sample motions within that shard according to
\begin{equation}
\label{eq:tracker_motion_sampling}
 P(i\mid g)=\frac{w_i}{\sum_{j\in\mathcal{S}_g} w_j}.
\end{equation}
Thus, repeated success reduces a motion's weight, while a failure restores its priority.
The weight floor preserves a nonzero sampling probability for every motion.

\subsection{Survival Rate and Termination Conditions}
\label{app:tracking_terminations}

The \emph{survival rate} reported in the results tables is the percentage of evaluation rollouts in which neither a \emph{robot fall} nor an \emph{object drop} occurs. As such, the survival rates for kicking and push/pull tasks are high because the object is always on the ground. \emph{Task success} separately measures whether the final object pose satisfies the goal tolerances. The full set of failure conditions below governs episode termination during tracking controller pretraining (Section~\ref{subsec:tracking_pretraining}) and evaluation-based motion reweighting (Appendix~\ref{app:tracking_training}):
\begin{enumerate}
    \item \emph{Robot fall:} the absolute height error of the torso or any wrist or ankle exceeds $0.25$\,m.
    \item \emph{Object drop:} the absolute object-height error exceeds $1$\,m.
    \item The absolute difference between the $z$-components of gravity projected into the actual and reference torso frames exceeds $0.8$.
    \item Any component of the base angular velocity exceeds $8.73$\,rad/s in magnitude.
\end{enumerate}
The object-height threshold follows a piecewise-linear schedule through $1.0$, $0.3$, $0.6$, and $0.3$\,m at iterations $5000$, $15{,}000$, $20{,}000$, and $25{,}000$, respectively, and is held fixed during each evaluation.
Reaching the end of the reference motion or the $10$\,s episode limit without a failure is treated as a timeout and counts towards the survival rate.

\subsection{Tracking Controller Reward Terms}
\label{app:tracking_rewards}

Table~\ref{tab:reward_terms} lists the reward terms used to pretrain the generalist tracking controller (Section~\ref{subsec:tracking_pretraining}).

\begin{table}[!h]
    \centering
    \caption{Reward terms used by the robot-object multi-motion tracking controller. }
    \label{tab:reward_terms}
    \small
    \setlength{\tabcolsep}{4pt}
    \renewcommand{\arraystretch}{1.1}
    \resizebox{\columnwidth}{!}{%
        \begin{tabular}{@{}lcc@{}}
            \toprule
            \textbf{Reward Term} & \textbf{Equation} & \textbf{Weight} \\
            \midrule
            \multicolumn{3}{l}{\textit{Motion Tracking}} \\
            \midrule
            Root Position & $\exp\!\left(-5\|\mathbf{p}_t - \mathbf{\hat{p}}_t^{\text{D}}\|^2\right)$ & 0.5 \\
            Root Orientation & $\exp\!\left(-3\|\mathbf{q}_t - \mathbf{\hat{q}}_t^{\text{D}}\|^2\right)$ & 0.5 \\
            Body Position & $\exp\!\left(-5\|\mathbf{p}_{\text{body},t} - \mathbf{\hat{p}}_{\text{body},t}^{\text{D}}\|^2\right)$ & 1.0 \\
            Body Orientation & $\exp\!\left(-3\|\boldsymbol{\theta}_{\text{body},t} - \boldsymbol{\hat{\theta}}_{\text{body},t}^{\text{D}}\|^2\right)$ & 1.0 \\
            Body Linear Velocity & $\exp\!\left(-0.5\|\mathbf{v}_t - \mathbf{\hat{v}}_t^{\text{D}}\|^2\right)$ & 1.0 \\
            Body Angular Velocity & $\exp\!\left(-0.05\|\boldsymbol{\omega}_t - \boldsymbol{\hat{\omega}}_t^{\text{D}}\|^2\right)$ & 1.0 \\
            Joint Pos Tracking & $\exp\!\left(-5\|\boldsymbol{u}_t - \boldsymbol{\hat{u}}_t^{\text{D}}\|^2\right)$ & 2.0 \\
            \midrule
            \multicolumn{3}{l}{\textit{Object Tracking}} \\
            \midrule
            Contact Match & \scalebox{0.85}{$\mathbb{I}(c = \hat{c}^{\text{D}})\, \cdot \exp\!\left(-0.1\left(\left\|\boldsymbol{F}_t\right\| - 10\right)\right)$} & 1.25 \\
            Object Position & $\exp\!\left(-8\|\mathbf{p}_{\text{obj},t} - \mathbf{\hat{p}}_{\text{obj},t}^{\text{D}}\|^2\right)$ & 1.0 \\
            Object Orientation & $\exp\!\left(-5\|\boldsymbol{\theta}_{\text{obj},t} - \boldsymbol{\hat{\theta}}_{\text{obj},t}^{\text{D}}\|^2\right)$ & 1.0 \\
            Object Linear Velocity & $\exp\!\left(-2\|\mathbf{v}_{\text{obj},t} - \mathbf{\hat{v}}_{\text{obj},t}^{\text{D}}\|^2\right)$ & 1.0 \\
            Object Angular Velocity & $\exp\!\left(-0.2\|\boldsymbol{\omega}_{\text{obj},t} - \boldsymbol{\hat{\omega}}_{\text{obj},t}^{\text{D}}\|^2\right)$ & 1.0 \\
            \midrule
            \multicolumn{3}{l}{\textit{Regularization}} \\
            \midrule
            Action Rate & $-\|\mathbf{a}_t - \mathbf{a}_{t-1}\|^2$ & $0.1$ \\
            Joint Limit & $-\sum_i \phi(q_i, q_{i,\min}, q_{i,\max})$ & $10.0$ \\
            Self-collisions & $\sum_{c \in \mathcal{C}_{\text{self}}} 1$ & $-1.0$ \\
            \bottomrule
        \end{tabular}}
\end{table}

\subsection{Post-Training Reward Terms}
\label{app:posttraining_rewards}

Table~\ref{tab:pt_reward_terms} lists the reward terms used to finetune the diffusion motion generator (Section~\ref{subsubsec:da_ppo}).

\textit{Tracker fine-tuning rewards.} The tracker fine-tuning baselines of Table~\ref{tab:tracker_ft} (Section~\ref{subsec:res_tracker_ft}) compares tracker finetuning against different sets of rewards: $r_\text{kin}$, $r_\text{kin} + r_\text{goal}$, and $r_\text{gen}$.
$r_\text{gen}$ is a tracking reward of the same form as $r_\text{kin}$, but it measures the error of the realised motion against the reference generated by the diffusion motion generator, $\hat{s}^{D}_{\tau}$, which is the motion the tracker is commanded to follow, instead of against the IK-retargeted motion $s^{IK}_{\tau}$.

\begin{table}[!h]
    \centering
    \caption{Post-training reward terms for the diffusion motion generator over a chunk
    of $T$ steps.
    $(\hat{\cdot})$: realised motion; $(\cdot)^{IK}$: IK-retargeted motion; $(\cdot)^G$: goal object pose the generator is conditioned on. All squared-norm errors below are averaged over the chunk: $e \equiv \tfrac{1}{T}\sum_{t=1}^{T} e_t$.
    In the goal terms, $\theta_t$ is the angle between the realised and goal object orientations, with $\sigma_p = 7.8$\,cm and $\sigma_\theta = 0.244$\,rad.
    }
    \label{tab:pt_reward_terms}
    \small
    \setlength{\tabcolsep}{1pt}
    \renewcommand{\arraystretch}{0.5}
    \resizebox{\columnwidth}{!}{%
        \begin{tabular}{@{}lcc@{}}
            \toprule
            \textbf{Reward Set} & \textbf{Equation} & \textbf{Weight} \\
            \midrule
            \multicolumn{3}{l}{\textit{Kinematic Fidelity ($r_\text{kin}$)}} \\
            \midrule
            Anchor Position & $\exp(-5\overline{\|\mathbf{\hat{p}}_t - \mathbf{p}_t^{IK}\|^2})$ & 0.25 \\
            Anchor Orientation & $\exp(-3\overline{\|\mathbf{\hat{q}}_t - \mathbf{q}_t^{IK}\|^2})$ & 0.25 \\
            Body Position & $\exp(-5\overline{\|\mathbf{\hat{p}}_{\text{body},t} - \mathbf{p}_{\text{body},t}^{IK}\|^2})$ & 0.50 \\
            Body Orientation & $\exp(-3\overline{\|\mathbf{\hat{q}}_{\text{body},t} - \mathbf{q}_{\text{body},t}^{IK}\|^2})$ & 0.50 \\
            Object Position & $\exp(-8\overline{\|\mathbf{\hat{p}}_{\text{obj},t} - \mathbf{p}_{\text{obj},t}^{IK}\|^2})$ & 0.25 \\
            Object Orientation & $\exp(-5\overline{\|\mathbf{\hat{q}}_{\text{obj},t} - \mathbf{q}_{\text{obj},t}^{IK}\|^2})$ & 0.25 \\
            Termination & $\mathbf{1}[\text{terminated during chunk}]$ & $-5.0$ \\
            \midrule
            \multicolumn{3}{l}{\textit{Goal ($r_\text{goal}$)}} \\
            \midrule
            Object Goal Pose & $\exp(-\overline{\|\mathbf{\hat{p}}_{\text{obj},t} - \mathbf{p}_{\text{obj}}^{G}\|^2}/\sigma_p^2 - \overline{\theta_t^2}/\sigma_\theta^2)$ & 5.00 \\
            Survival & $\tfrac{1}{T}\sum_{t=1}^{T} \mathbf{1}[\text{alive at step } t]$ & 1.00 \\
            
            \bottomrule
        \end{tabular}}
\end{table}

\subsection{DA:PPO Post-Training Hyperparameters}
\label{app:dappo_hparams}

Table~\ref{tab:dappo_hparams} lists the configuration used to post-train the diffusion motion generator with DA:PPO (Section~\ref{subsubsec:da_ppo}).
The DPPO-style baseline uses the same configuration with $\mathcal{T} = 0$ and fixed step variances.

\begin{table}[!h]
    \centering
    \caption{DA:PPO post-training configuration of the diffusion motion generator.}
    \label{tab:dappo_hparams}
    \small
    \setlength{\tabcolsep}{4pt}
    \renewcommand{\arraystretch}{1.1}
    \begin{tabularx}{\columnwidth}{@{}Xl@{}}
        \toprule
        \textbf{Parameter} & \textbf{Value} \\
        \midrule
        \multicolumn{2}{l}{\textit{DA-MDP}} \\
        \midrule
        Chunk horizon $n$ / executed frames $m$ & $64$ / $56$ \\
        Denoising steps finetuned & all $K = 10$ \\
        Temperature $\mathcal{T}$ & $10^{-4}$ \\
        Discount $\gamma$ / GAE $\lambda$ & $1.0$ / $0.95$ \\
        Step-variance multipliers & learned, initialized at $1$ \\
        \quad shared with the forward kernel & yes \\
        Reward normalization & running return std, clip $\pm 10$ \\
        \midrule
        \multicolumn{2}{l}{\textit{DA:PPO update}} \\
        \midrule
        Clipping range $\epsilon$ & $0.2$ \\
        Epochs per iteration & $2$ \\
        Minibatch size (denoising transitions) & $1024$ \\
        Chunks per environment per iteration & $16$ \\
        Iterations & $500$ \\
        Parallel environments & $2048$ \\
        Advantage normalization & per rollout \\
        Gradient-norm clip & $1.0$ \\
        \midrule
        \multicolumn{2}{l}{\textit{Optimization}} \\
        \midrule
        Optimizer & AdamW \\
        Learning rate, generator / step variances & $10^{-6}$ / $10^{-6}$ \\
        Learning rate, critic & $10^{-4}$ \\
        \midrule
        \multicolumn{2}{l}{\textit{Critic}} \\
        \midrule
        Architecture & residual MLP, $3 \times 256$, Mish \\
        Value-loss coefficient / clip & $1.0$ / $0.2$ \\
        Critic-only warm-up & $10$ iterations $\times$ $4$ epochs \\
        \bottomrule
    \end{tabularx}
\end{table}
\subsection{Tracker Fine-Tuning Hyperparameters}
\label{app:kl_lora}

Table~\ref{tab:tracker_ft_hparams} lists the configuration used to fine-tune the tracking policy in Table~\ref{tab:tracker_ft} (Section~\ref{subsec:res_tracker_ft}).
The motion generator stays frozen, and the rollouts are collected in the same closed loop as the evaluation; the reward sets are those of Appendix~\ref{app:posttraining_rewards}.

\begin{table}[!h]
    \centering
    \caption{Configuration of tracker fine-tuning with PPO.}
    \label{tab:tracker_ft_hparams}
    \small
    \setlength{\tabcolsep}{4pt}
    \renewcommand{\arraystretch}{1.1}
    \begin{tabularx}{\columnwidth}{@{}Xl@{}}
        \toprule
        \textbf{Parameter} & \textbf{Value} \\
        \midrule
        \multicolumn{2}{l}{\textit{Rollouts}} \\
        \midrule
        Executed frames per re-plan $m$ & $56$ \\
        Parallel environments & $2048$ \\
        Iterations & $1000$ \\
        \midrule
        \multicolumn{2}{l}{\textit{PPO update}} \\
        \midrule
        Discount $\gamma$ / GAE $\lambda$ & $0.99$ / $0.95$ \\
        Clipping range $\epsilon$ & $0.2$ \\
        Epochs per iteration & $5$ \\
        Minibatch size (frames) & $16{,}384$ \\
        Early stop of the epochs (approx.\ KL) & $0.02$ \\
        Entropy bonus & $0$ \\
        Gradient-norm clip & $1.0$ \\
        \midrule
        \multicolumn{2}{l}{\textit{Critic}} \\
        \midrule
        Initialization & pretrained asymmetric critic \\
        Learning rate & $10^{-3}$ \\
        \bottomrule
    \end{tabularx}
\end{table}

\subsection{SD:DPO Objective and its Weighting}
\label{app:sdpo_derivation}

SD:DPO adapts the objective of SDPO~\cite{hubotter2026reinforcement}.
For a prompt $x$ and a response $y$ sampled from the student, SDPO minimises
\begin{equation*}
\mathcal L_{\mathrm{SDPO}}(\theta)
= \sum_t \mathrm{KL}\big(\pi_\theta(\cdot\mid x,y_{<t})\,\big\|\,\operatorname{sg}\bigl(\pi_\theta(\cdot\mid x,f,y_{<t})\bigr)\big),
\end{equation*}
the KL divergence at every position $t$ between the student's next-token distribution and that of a self-teacher, the same model additionally conditioned on feedback $f$.
Mapping tokens to denoising steps, prefixes to noisy chunks with their context, next-token distributions to reverse kernels, and the feedback $f$ to the kinematic reference $s^{IK}_\tau$, and sampling $\hat s^{D,k}_\tau$ and $k$ as in~\eqref{eq:sdpo_sampling}, gives
\begin{equation}
\label{eq:sdpo_diffusion}
\mathcal L_{\mathrm{KL}}(\phi)
= \mathbb E_{\mathcal B,k,q}\Big[\mathrm{KL}\big(p^{\varnothing}_\phi\,\big\|\,\operatorname{sg}(p^{IK}_\phi)\big)\Big],
\end{equation}
where $p^{\varnothing}_\phi = p_\phi(\cdot\mid\hat s^{D,k}_\tau,c_t,\varnothing)$ and $p^{IK}_\phi = p_\phi(\cdot\mid\hat s^{D,k}_\tau,c_t,s^{IK}_\tau)$ are the student's and the teacher's reverse kernels over $(\hat x^{k-1},\hat R^{k-1})$ given by~\eqref{eq:euc_reverse} and~\eqref{eq:so3_reverse}.
As in SDPO, the stop-gradient keeps the teacher from drifting towards the student, and in practice the teacher is evaluated with the exponential moving average weights $\bar\phi$, as in~\eqref{eq:sdpo_practice}.
Because the Euclidean and rotational parts of each kernel are sampled independently given $\hat s^{D,k}_\tau$, the KL splits into a Euclidean and a rotational term.
The remainder of this appendix relates~\eqref{eq:sdpo_diffusion} to the regression of~\eqref{eq:sdpo_practice} that is minimised in practice.
At denoising step $k$, we use a noisy motion $\hat{s}^{D,k}_{\tau}$ obtained via the forward kernel from the student's rollout in closed-loop, where $\hat{x}^{k}$ corresponds to the corresponding noisy Euclidean features.
The student and the teacher evaluate the same network at the same $(\hat{s}^{D,k}_{\tau}, c,  k)$ and differ only in the prior token ($\varnothing$ or $s^{IK}_{\tau}$); we mark their reverse kernels of $\hat{x}^{k-1}$, their $\mathbf{v}$-predictions and their clean estimates with the corresponding superscript.
By~\eqref{eq:euc_reverse} both kernels are Gaussian with covariance $\tilde{\beta}_k I$ and mean $a_k\,\hat{x}^{0|k} + b_k\,\hat{x}^{k}$, and $\hat{x}^{0|k} = \sqrt{\bar{\alpha}_k}\,\hat{x}^{k} - \sqrt{1-\bar{\alpha}_k}\,\hat{\mathbf{v}}^{k}$, so that
\begin{equation}
\label{eq:sdpo_kl_weight}
\begin{aligned}
\mathrm{KL}\big(p^{\varnothing}_\phi \,\big\|\, p^{IK}_\phi\big)
&= \frac{a_k^2}{2\tilde{\beta}_k}\, \big\| \hat{x}^{0|k,\varnothing} - \hat{x}^{0|k,IK} \big\|^2 \\
&= w_k\, \big\| \hat{\mathbf{v}}^{k,\varnothing} - \hat{\mathbf{v}}^{k,IK} \big\|^2, \\
w_k &= \frac{a_k^2\,(1-\bar{\alpha}_k)}{2\tilde{\beta}_k} = \frac{\beta_k}{2}\,\frac{\bar{\alpha}_{k-1}}{1-\bar{\alpha}_{k-1}} .
\end{aligned}
\end{equation}
The KL is therefore the squared difference of the two network predictions, scaled by a weight that depends only on the diffusion step.
The same holds for the rotational features to first order in the tangent space, with the tangent-noise predictions in place of $\hat{\mathbf{v}}^{k}$ and another step-dependent weight.

We set $w_k = 1$ for two reasons.
First, $w_k$ cannot be used as it stands. Since $\bar{\alpha}_0 = 1$, it is infinite at $k = 1$, where the reverse kernel is deterministic.
Over the remaining steps in our schedule, the exact weighting would also concentrate the loss on the least-noisy steps.
Second, the weights do not change what is learned. For any positive weights, the objective is non-negative and vanishes only when the student's prediction equals the teacher's at every step, which is also where~\eqref{eq:sdpo_diffusion} attains its minimum; the weights only decide how errors at different noise levels are traded off before that point is reached.
Equal weights are the choice of the simplified diffusion objective~\cite{ho2020denoisingdiffusionprobabilisticmodels}, and they are applied here in the space the network regresses in, where the $\mathbf{v}$-target has a comparable scale at every noise level~\cite{salimans2022progressive}.
Setting $w_k = 1$ for the Euclidean and the rotational features gives the distillation term of~\eqref{eq:sdpo_practice}.

\subsection{Action-Level Diffusion Baselines}
\label{app:action_baselines}

This appendix describes the action-level baselines of Section~\ref{subsec:res_action}: an action-only diffusion policy in the style of PDP~\cite{truong2024pdp}, and a state--action diffusion model in the style of BeyondMimic~\cite{liao2025beyondmimicmotiontrackingversatile}.
Both output robot joint PD targets directly, with no tracker in the loop.
Table~\ref{tab:action_baseline_hparams} lists their configuration.

\textit{Training data.}
Both baselines are trained via behaviour cloning on rollouts of our pretrained generalist tracking policy over every motion of the SBTO- and symmetry-augmented data corpus, under the randomisation of Table~\ref{tab:dr1}.
As in PDP~\cite{truong2024pdp}, the joint targets applied to the robot are perturbed using temporally correlated Ornstein--Uhlenbeck noise as used in~\cite{liao2025beyondmimicmotiontrackingversatile}, while the tracker's unperturbed output is stored as the label, so that the data pairs off-nominal states with the actions that correct them.
We collect full-length rollouts and $5$\,s rollouts started from random frames of each motion, and discard every rollout that fails, generating $15$M frames ($85$\,h) of state--action data.

\textit{Models.}
The state-action diffusion baseline uses the diffusion transformer (DiT) architecture of~\cite{Huang_2025}, for a faithful reproduction of the results.
Here, a DiT learns to denoise a chunk of joint state--action tokens, each with its own noise level, given the current state and a history of state--action pairs; state tokens attend to all states and to no action, and action tokens attend only to states and actions up to their own step (causal attention).
PDP~\cite{truong2024pdp} only learns to diffuse the action distribution.
For the PDP-style baseline, we therefore drop the state prediction from the same network and exclude it from the loss, leaving an action diffusion policy conditioned on the state--action history and the task.
Both models replan at every control step ($50$\,Hz) with $20$ denoising steps and execute only the first action.

The network, the diffusion process, the optimizer and the collection of expert rollouts follow Diffuse-CLoC~\cite{Huang_2025} and the first version of BeyondMimic~\cite{liao2025beyondmimicmotiontrackingversatile}, which applies Diffuse-CLoC to the robot we use.
Where the two differ (attention heads, horizons, state features) we take the BeyondMimic values.
We depart from both works in the following places:
\begin{enumerate}
    \item The state additionally contains the object's observations such as its position, orientation and linear velocity.
    \item The conditioned models receive the goal and the IK-retargeted motion as additional context tokens.
    \item Later versions of BeyondMimic diffuse the latent of an action autoencoder. We diffuse joint targets directly, as Diffuse-CLoC and the first version of BeyondMimic do.
\end{enumerate}

\textit{Unconditioned prior.}
In~\cite{Huang_2025,liao2025beyondmimicmotiontrackingversatile} the state-action diffusion model is not conditioned on the task: it is an unconditioned prior that is steered at inference using a task-specific guidance cost.
Hence, we train a separate diffusion model without the conditioning tokens and evaluate it at test time by steering with the IK-retargeted motion.
Unlike the conditioned models, it takes its horizons and its sampling of noise levels from Diffuse-CLoC; these levels include zero, so that the clean keyframes of in-betweening are seen in training.

\textit{Test-time steering.}
The guidance cost sums, over the predicted states of the horizon, the squared position error and the orientation error $1 - \langle q, q^{IK} \rangle^2$ of the root and of the object with respect to the IK-retargeted motion, with $q$ a unit quaternion.
At every denoising step the predicted clean states are moved against the gradient of this cost, scaled by the guidance weight.
Keyframe in-betweening instead inpaints a single state ahead in time: the keyframe is the state of the IK-retargeted motion at the given lookahead, expressed relative to the measured planar position and heading of the robot, and the waypoint cost of~\cite{liao2025beyondmimicmotiontrackingversatile} pulls the planar root position toward that of the keyframe.
The steered row of Table~\ref{tab:motion_vs_action_generator} uses in-betweening with the waypoint cost.
A lookahead of $0.32$\,s gave the highest survival rate among $0.08$, $0.24$, $0.32$, $0.64$ and $0.96$\,s.

\begin{table}[!h]
    \centering
    \caption{Configuration of the action-level diffusion baselines. Where two values are separated by a bar, the first holds for the two conditioned models and the second for the unconditioned prior.}
    \label{tab:action_baseline_hparams}
    \small
    \setlength{\tabcolsep}{4pt}
    \renewcommand{\arraystretch}{1.1}
    \begin{tabularx}{\columnwidth}{@{}Xl@{}}
        \toprule
        \textbf{Parameter} & \textbf{Value} \\
        \midrule
        \multicolumn{2}{l}{\textit{Expert rollouts}} \\
        \midrule
        Expert & Pretrained Tracking Policy \\
        Randomization & Table~\ref{tab:dr1} \\
        Noise on applied joint targets & Ornstein--Uhlenbeck, $\theta = 0.8$ \\
        \quad scale $\sigma$ (normalized action units) & $0.1$ and $0.25$ \\
        Stored action & tracker output without noise \\
        Rollout length & full motion, or $5$\,s \\
        Rollouts kept / frames & $59.8$k / $15.4$M \\
        \midrule
        \multicolumn{2}{l}{\textit{Representation}} \\
        \midrule
        Control and data rate & $50$\,Hz \\
        History & $4$ state--action pairs \\
        State / action horizon (steps) & $16$ / $8$ $\,|\,$ $48$ / $16$ \\
        State dimension & $192$ \\
        \quad root position, velocity, orientation & $9$ \\
        \quad body positions and velocities & $174$ \\
        \quad object pose and velocity & $9$ \\
        Action & $29$ joint position targets \\
        \midrule
        \multicolumn{2}{l}{\textit{Denoiser and diffusion process}} \\
        \midrule
        Layers / heads / width & $6$ / $4$ / $512$ \\
        Attention dropout & $0.3$ \\
        Trainable parameters & $20.2$M \\
        Denoising steps / schedule & $20$ / cosine \\
        Prediction target / loss & clean sample / MSE \\
        Training noise levels (per token) & $1$--$20$ $\,|\,$ $0$--$20$ \\
        Action-loss weight & $1$ $\,|\,$ $13$ \\
        Reverse sampler & DDPM \\
        \midrule
        \multicolumn{2}{l}{\textit{Optimization}} \\
        \midrule
        Optimizer & AdamW \\
        Learning rate / weight decay & $10^{-4}$ / $10^{-3}$ \\
        LR schedule & cosine, $2000$ warmup \\
        Batch size (windows) & $512$ \\
        Gradient-norm clip / EMA decay & $1.0$ / $0.999$ \\
        Gradient steps& $300$k \\
        \midrule
        \multicolumn{2}{l}{\textit{Test-time steering}} \\
        \midrule
        Guidance weight & $0.003$--$0.1$ \\
        Keyframe lookahead & $16$ steps ($0.32$\,s) \\
        Waypoint guidance weight & $10$ \\
        Rolling noise levels, states / actions & $14$ / $4$ \\
        \bottomrule
    \end{tabularx}
\end{table}

\subsection{Ablation on Diffusion Prediction Horizon}
\label{app:horizon}

This appendix examines how far into the future the diffusion motion generator can plan.
We pretrain five generators that differ only in the prediction horizon $n \in \{20, 32, 48, 64, 100\}$ frames, i.e.\ $0.4$\,s to $2.0$\,s of motion at $50$\,Hz.
The denoiser remains unchanged; only its input sequence grows, since the prior and the generated chunk each contribute $n$ tokens.
The tracker reads its reference up to $8$ frames ahead (Appendix~\ref{app:tracking_training}), so the last $8$ frames of a chunk are kept as its lookahead, and the first $m = n - 8$ frames are executed, as in our default of $n = 64$ and $m = 56$.
A longer horizon, therefore, also means that a longer segment is executed before the generator re-plans from the reached state: the number of generator calls per second of motion falls from $4.2$ at $n = 20$ to $0.54$ at $n = 100$.

All other settings follow Table~\ref{tab:pretrain_hparams}, and the absolute numbers below should not be compared with those of the main experiments.

\begin{table}[!h]
    \centering
    \caption{Closed-loop performance of pretrained diffusion motion generators that differ only in the prediction horizon $n$, with $m = n - 8$ frames executed per re-plan. Evaluated with the frozen tracker over $16$ rollouts on each motion. Values are mean $\pm$ SEM over motions.}
    \label{tab:horizon}
    \small
    \setlength{\tabcolsep}{4pt}
    \renewcommand{\arraystretch}{1.1}
    \resizebox{\columnwidth}{!}{%
        \begin{tabular}{@{}rrccccc@{}}
            \toprule
            $\boldsymbol{n}$ & $\boldsymbol{m}$ & \textbf{Survival} & \textbf{Task success} & \textbf{MPKPE} & \textbf{Obj. pos.} & \textbf{Obj. ori.} \\
            & & \textbf{(\%) $\uparrow$} & \textbf{(\%) $\uparrow$} & \textbf{(cm) $\downarrow$} & \textbf{(cm) $\downarrow$} & \textbf{($^\circ$) $\downarrow$} \\
            \midrule
            \multicolumn{7}{@{}l}{\textit{Without randomization}} \\
            \midrule
            $20$ & $12$ & 97.6 $\pm$ 2.3 & 89.7 $\pm$ 5.1 & 5.9 $\pm$ 0.4 & 4.6 $\pm$ 1.3 & 9.0 $\pm$ 2.7 \\
            $32$ & $24$ & 100.0 $\pm$ 0.0 & 92.9 $\pm$ 4.7 & 5.6 $\pm$ 0.3 & 3.6 $\pm$ 0.4 & 6.6 $\pm$ 1.7 \\
            $48$ & $40$ & 97.4 $\pm$ 2.5 & 87.7 $\pm$ 5.5 & 5.8 $\pm$ 0.5 & 5.2 $\pm$ 1.8 & 8.9 $\pm$ 3.0 \\
            $64$ & $56$ & 100.0 $\pm$ 0.0 & 93.1 $\pm$ 4.7 & 5.0 $\pm$ 0.2 & 3.4 $\pm$ 0.4 & 7.0 $\pm$ 1.8 \\
            $100$ & $92$ & 100.0 $\pm$ 0.0 & 91.8 $\pm$ 4.7 & 5.1 $\pm$ 0.3 & 3.2 $\pm$ 0.4 & 6.0 $\pm$ 1.8 \\
            \midrule
            \multicolumn{7}{@{}l}{\textit{With the randomization of Table~\ref{tab:dr1}}} \\
            \midrule
            $20$ & $12$ & 99.6 $\pm$ 0.3 & 90.5 $\pm$ 4.5 & 6.6 $\pm$ 0.3 & 4.9 $\pm$ 0.5 & 9.4 $\pm$ 1.8 \\
            $32$ & $24$ & 99.1 $\pm$ 0.8 & 92.0 $\pm$ 3.8 & 6.7 $\pm$ 0.3 & 5.2 $\pm$ 0.7 & 9.5 $\pm$ 1.8 \\
            $48$ & $40$ & 99.6 $\pm$ 0.4 & 89.9 $\pm$ 4.9 & 6.4 $\pm$ 0.3 & 4.9 $\pm$ 0.6 & 9.7 $\pm$ 2.1 \\
            $64$ & $56$ & 100.0 $\pm$ 0.0 & 91.8 $\pm$ 4.4 & 6.2 $\pm$ 0.2 & 4.6 $\pm$ 0.4 & 8.9 $\pm$ 1.7 \\
            $100$ & $92$ & 99.6 $\pm$ 0.4 & 91.4 $\pm$ 4.1 & 6.3 $\pm$ 0.3 & 4.6 $\pm$ 0.5 & 8.8 $\pm$ 1.8 \\
            \bottomrule
        \end{tabular}}
\end{table}

Table~\ref{tab:horizon} reports the closed-loop performance with the frozen tracker.
Predicting up to $2.0$\,s and executing up to $1.84$\,s per re-plan does not measurably degrade any metric in simulation.
The survival rate remains above $97\%$ for every horizon, and task success varies by at most $5.4$ points across horizons, which is about one SEM, with no trend in $n$.
The tracking errors do not grow with the horizon either: MPKPE and object position error are lowest for $n = 64$ and $n = 100$ in both settings.

\subsection{Goal Generalization Before DA:PPO Finetuning}
\label{app:goal_reach_maps}

\begin{figure}[t]
    \centering
    \begin{subfigure}{\columnwidth}
        \centering
        \includegraphics[width=\columnwidth]{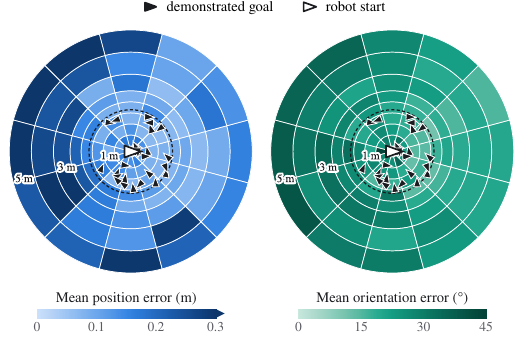}
        \caption{Student (pretrained)}
        \label{fig:goal_reach_map_student}
    \end{subfigure}

    \vspace{6pt}
    \begin{subfigure}{\columnwidth}
        \centering
        \includegraphics[width=\columnwidth]{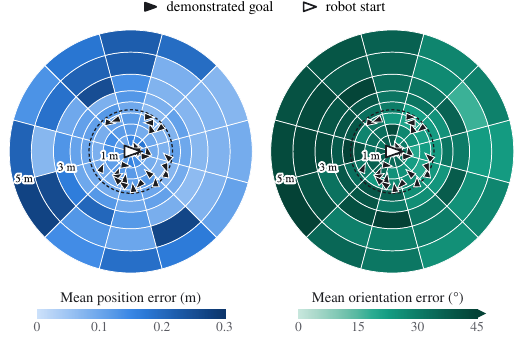}
        \caption{Student $+$ SD:DPO}
        \label{fig:goal_reach_map_sdpo}
    \end{subfigure}
    \caption{Goal grid for (a)~the pretrained Student and (b)~Student $+$ SD:DPO, on the same episodes and color scales. Each map shows the mean final position error (left) and orientation error (right) of the object over the rollouts in which it was lifted by more than $12$\,cm: $83\%$ in (a) and $78\%$ in (b), against $90\%$ for the final checkpoint. An arrow at the end of a color bar marks goal positions whose mean lies beyond its range, up to $0.48$\,m in (a) and $46.6^\circ$ in (b).}
    \label{fig:goal_reach_map_earlier}
\end{figure}

Fig.~\ref{fig:goal_reach_map_earlier} repeats the goal-grid evaluation of Sec.~\ref{subsec:res_sdpo} for the two checkpoints that precede the final one in Table~\ref{tab:sdpo_tasks}: the pretrained \emph{Student} and \emph{Student $+$ SD:DPO}, from which DA:PPO starts.
All three checkpoints are evaluated on the same episodes, i.e.\ the same goal poses, initial object poses and evaluation seed, and the plots share their color scales.

The pretrained student has the highest position errors since the diffusion sampling is ill-conditioned. Self-distillation mitigates the position error a bit, but still struggles with object orientations. However, as shown in Fig.~\ref{fig:goal_reach_map}, the finetuning on top of the distilled student unlocks the final leg of performance, pushing diffusion sampling towards high reward samples.

\begin{figure*}[t]
      \centering
      \includegraphics[width=0.9\textwidth]{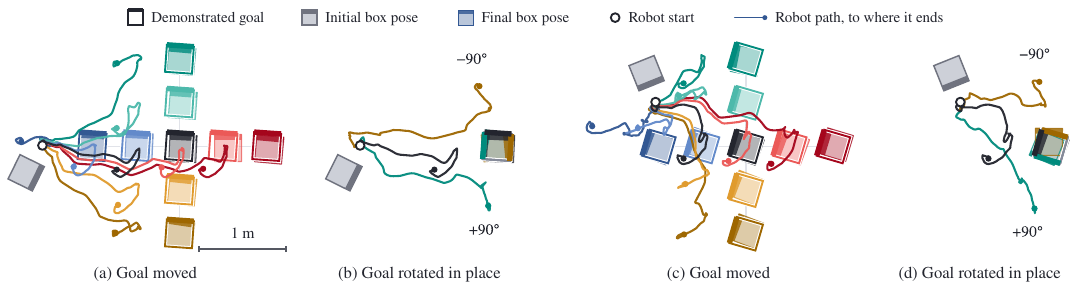}
      \caption{
      Robot $x-y$ trajectories for new goals chosen outside the demonstration set, for two different motions (a, b and c, d). In (a,c), we displace the goal by $0.5$m-$1.0$m along $x$ and $y$ directions, while in (b,d), we rotate the goal pose around $z$ axis by $\pm90\deg$. We use different colors for new goals chosen away from the goal in the demonstration goal. The plots were generated using the Student $+$ SD:DPO $+$ DA:PPO model.}
      \label{fig:goal_generalization}
  \end{figure*}

In Fig.~\ref{fig:goal_generalization}, we vary the goal position from our demonstration by $\pm0.5$\,m and $\pm1$\,m along the $x$ and $y$ axes (a,c) and rotate it in place by $\pm 90^\circ$ (b,d), and roll out our final post-trained checkpoint, which is crucially only conditioned on the state-history and goal. We find that our self-distilled model can generate consistent motions to achieve the desired object goal pose.

\end{document}